\documentclass[letterpaper]{article} % DO NOT CHANGE THIS
\usepackage[preprint]{aaai2027} % DO NOT CHANGE THIS
\usepackage[hyphens]{url} % DO NOT CHANGE THIS
\usepackage{graphicx} % DO NOT CHANGE THIS
\usepackage{natbib} % DO NOT CHANGE THIS AND DO NOT ADD OPTIONS
\usepackage{caption} % DO NOT CHANGE THIS AND DO NOT ADD OPTIONS
\usepackage{amsmath}
\usepackage{amssymb}
\usepackage{booktabs}

\usepackage{arydshln}
\usepackage{listings}
\usepackage{tcolorbox}
\tcbuselibrary{skins,listings,breakable}
\newcommand{\spacedhdashline}{\noalign{\vskip 2pt}\hdashline\noalign{\vskip 2pt}}

\newtcblisting{appendixpromptbox}[1]{
  listing only, breakable, width=\columnwidth, colback=white,
  colframe=black, enhanced, sharp corners, boxrule=0.8pt, drop shadow,
  colbacktitle=black, coltitle=white, fonttitle=\bfseries\footnotesize,
  title=\textbf{#1}, left=0.8mm, right=0.8mm, top=0.5mm, bottom=0.5mm,
  before skip=5pt, after skip=7pt,
  listing options={basicstyle=\ttfamily\fontsize{6.5}{7.2}\selectfont,
    breaklines=true, breakatwhitespace=true, columns=fullflexible,
    keepspaces=true, showstringspaces=false}
}

\title{TextRefine: Improving Textual Fidelity, Spatial Placement, and Glyph Rendering for Text Editing in Product Posters}
\author{
Honglie Wang\textsuperscript{\rm 1,2,3},
Jia Sun\textsuperscript{\rm 1},
Zijun Li\textsuperscript{\rm 1},
Junlong Wu\textsuperscript{\rm 1},
Pengcheng Wei\textsuperscript{\rm 1},\\
Jiyuan Wang\textsuperscript{\rm 1},
Yongrui Heng\textsuperscript{\rm 1},
Boheng Zhang\textsuperscript{\rm 1},
Huaiqing Wang\textsuperscript{\rm 1},
Dewen Fan\textsuperscript{\rm 1},
Qianqian Gan\textsuperscript{\rm 1},
Fan Yang\textsuperscript{\rm 1},
Tingting Gao\textsuperscript{\rm 1},
Yan-Ming Zhang\textsuperscript{\rm 2,3}
}
\affiliations{
\textsuperscript{\rm 1}Kuaishou Technology\\
\textsuperscript{\rm 2}School of Artificial Intelligence, University of Chinese Academy of Sciences, Beijing 100049, China\\
\textsuperscript{\rm 3}State Key Laboratory of Multimodal Artificial Intelligence Systems (MAIS), Institute of Automation, Chinese Academy of Sciences, Beijing 100190, China
}

\begin{document}

\maketitle

\begin{abstract}
Text editing in product posters entails inserting new text or replacing existing text while preserving product appearance, background content, and global composition. Despite recent progress in instruction-based image editing, general-purpose models remain unreliable in this setting: they often omit or incorrectly render the target text, place it over salient products or pre-existing content, and produce structurally distorted or visually inconsistent glyphs. We introduce \textbf{TextRefine}, a task-aligned post-training framework that combines supervised fine-tuning with operation-specific reward optimization to address these complementary failure modes. For text insertion, our text-span-level reward jointly assesses semantic fidelity and target-span coverage, penalizes spatial conflicts with products and existing text, and employs a gated structural constraint to preserve non-text regions. For text replacement, our glyph-level reward leverages the connectionist temporal classification (CTC) posterior of the target character to provide graded supervision for fine-grained defects, including missing strokes, structural deformations, and confusion among visually similar characters. We further introduce \textbf{OpenTextEdit}, a dataset comprising 100K images for text editing in product posters, with multi-text layouts, detailed text attributes, product masks, and challenging low-frequency characters. Extensive experiments on both insertion and replacement demonstrate that TextRefine consistently outperforms the evaluated image editing baselines in textual fidelity, placement reliability, and glyph quality while better preserving source-image content.
\end{abstract}

\section{Introduction}
\label{sec:intro}

Text-to-image generation has advanced rapidly in recent years, and state-of-the-art models can now synthesize photorealistic, high-fidelity images from natural language prompts~\cite{SDXL,dalle3,imagen,flux,SD3.5}. Within this broader progress, Visual Text Rendering (VTR), the ability to generate legible and semantically faithful text inside images, has become a core capability of modern generative systems~\cite{aestheticsIsCheap,QwenImage,Seeddream4.0,team2026firered}. Recent specialized models already demonstrate strong text-rendering performance in the \emph{generation-from-scratch} regime, where dense multilingual text can be produced with high accuracy and visual coherence~\cite{anytext2,Glyph-byt5-v2,seedream2.0,Seedream3.0,QwenImage}.

\begin{figure}[t]
  \centering
  \includegraphics[width=0.8\linewidth]{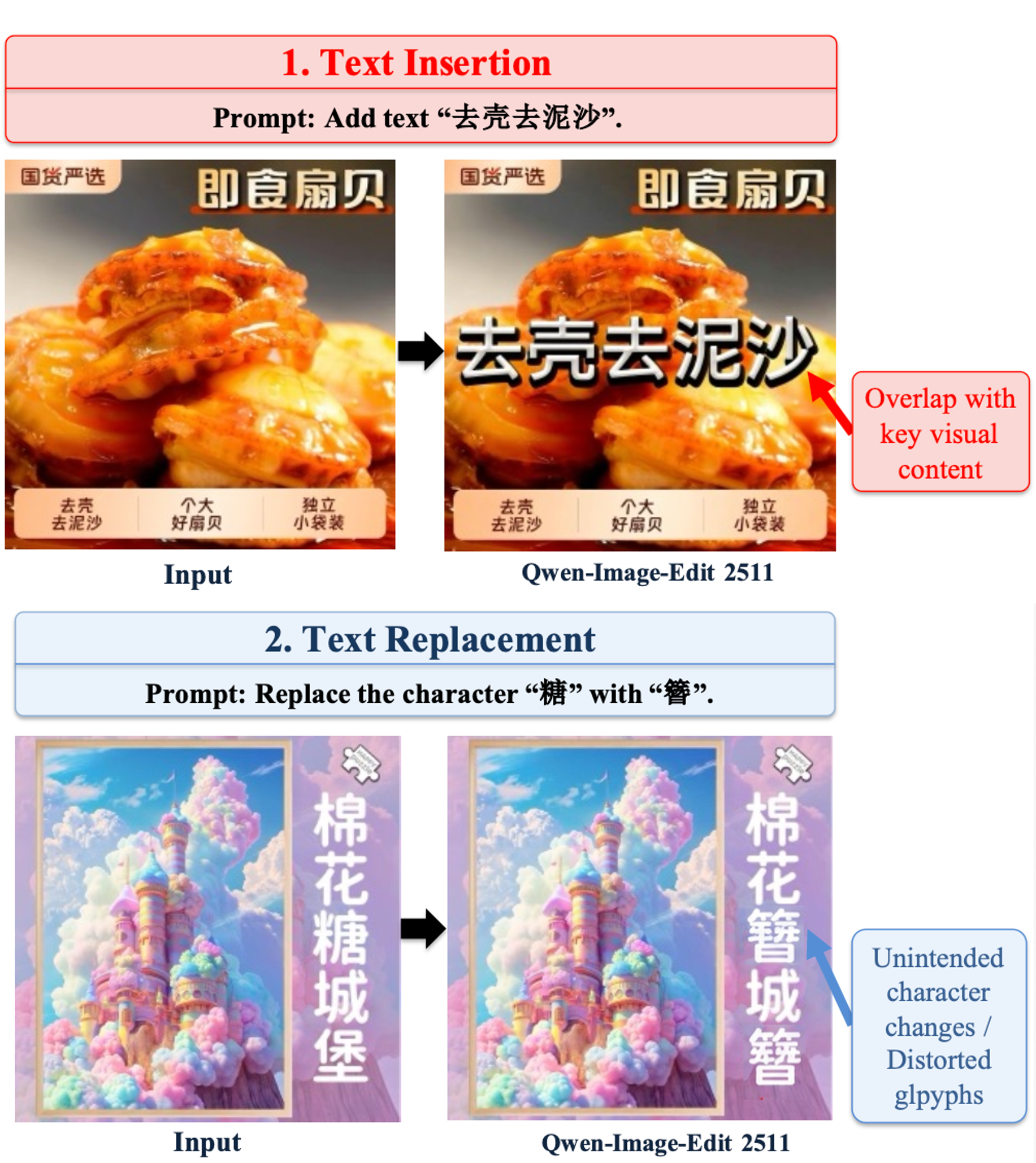}
  \caption{Representative failure modes of existing methods for text editing in product posters. In text insertion (top), generated text may overlap salient product regions, conflict with existing text, or deviate from the intended content. In text replacement (bottom), models may alter unintended character instances or produce structurally distorted glyphs.}
  \label{fig:teaser}
\end{figure}

Text editing in product posters presents challenges that are not adequately addressed by progress in generation from scratch. The task encompasses both \emph{text insertion}, in which one or more promotional text spans must be incorporated into an existing layout, and \emph{text replacement}, in which localized text must be modified without disrupting its surroundings. Successful editing therefore requires simultaneous control over three aspects: textual fidelity, spatial placement, and glyph rendering. As shown in Figure~\ref{fig:teaser}, existing models may omit or incorrectly render requested text, place newly generated text over salient products or pre-existing content, and introduce structural defects during character replacement. These errors are particularly consequential in product posters, where text conveys essential commercial information and must coexist with tightly constrained visual layouts. Moreover, all edits must preserve product appearance, background content, and the overall composition of the source image.

A central obstacle is the mismatch between these requirements and existing supervision. Reinforcement-learning approaches to visual text rendering commonly derive rewards from OCR or multimodal large language models and reduce recognition outputs to rule-based scores such as exact match or edit distance~\cite{X-Omni_RL_and_LongText,PPOCRV5,got_ocr2.0,Qwen2.5-VL,TextCrafter_cvtg2k,chang2025oneig}. Although such signals measure whether textual content is recognizable, they provide limited supervision for poster-specific spatial conflicts and are often insensitive to subtle glyph defects that do not change the recognized character. A single string-level score is therefore insufficient to capture the distinct requirements of insertion and replacement: insertion demands span-level assessment of content, coverage, and placement, whereas replacement requires fine-grained sensitivity to character structure.

To address this supervision gap, we propose \textbf{TextRefine}, a task-aligned post-training framework for text editing in product posters. TextRefine first performs supervised fine-tuning on mixed insertion and replacement data to establish general editing capability, and then applies reinforcement learning with two complementary, operation-specific reward signals. For text insertion, a text-span-level reward evaluates semantic fidelity and target-span coverage while penalizing conflicts with product regions and pre-existing text; a gated structural constraint additionally discourages unintended changes to non-text content. For text replacement, a glyph-level reward uses the connectionist temporal classification posterior of the target character to provide graded supervision for missing strokes, structural deformations, and confusion among visually similar characters.

We further construct \textbf{OpenTextEdit}, a 100K-image dataset designed for text editing in product posters. It covers both text insertion and text replacement, with an emphasis on multi-text layouts and structurally complex low-frequency characters, and provides text content, layout references, visual attributes, and product masks for task-aligned training and evaluation.

We summarize our contributions as follows:
\begin{itemize}
\item We formulate text editing in product posters as two complementary tasks---text insertion and text replacement---and identify the distinct requirements of textual fidelity, spatial placement, glyph rendering, and source-image preservation.
\item We propose TextRefine, a task-aligned post-training framework that combines supervised fine-tuning with reinforcement learning. Its text-span-level reward jointly optimizes content, coverage, and product-aware placement constraints for insertion, while its CTC-posterior-based glyph-level reward provides fine-grained structural supervision for replacement.
\item We construct OpenTextEdit, a 100K-image dataset featuring multi-text layouts, detailed text attributes, product masks, and challenging low-frequency characters. Experiments on both tasks demonstrate consistent improvements over strong image editing baselines in text accuracy, placement reliability, glyph fidelity, and content preservation.
\end{itemize}

\begin{figure*}[t]
  \centering
  \includegraphics[width=0.9\textwidth]{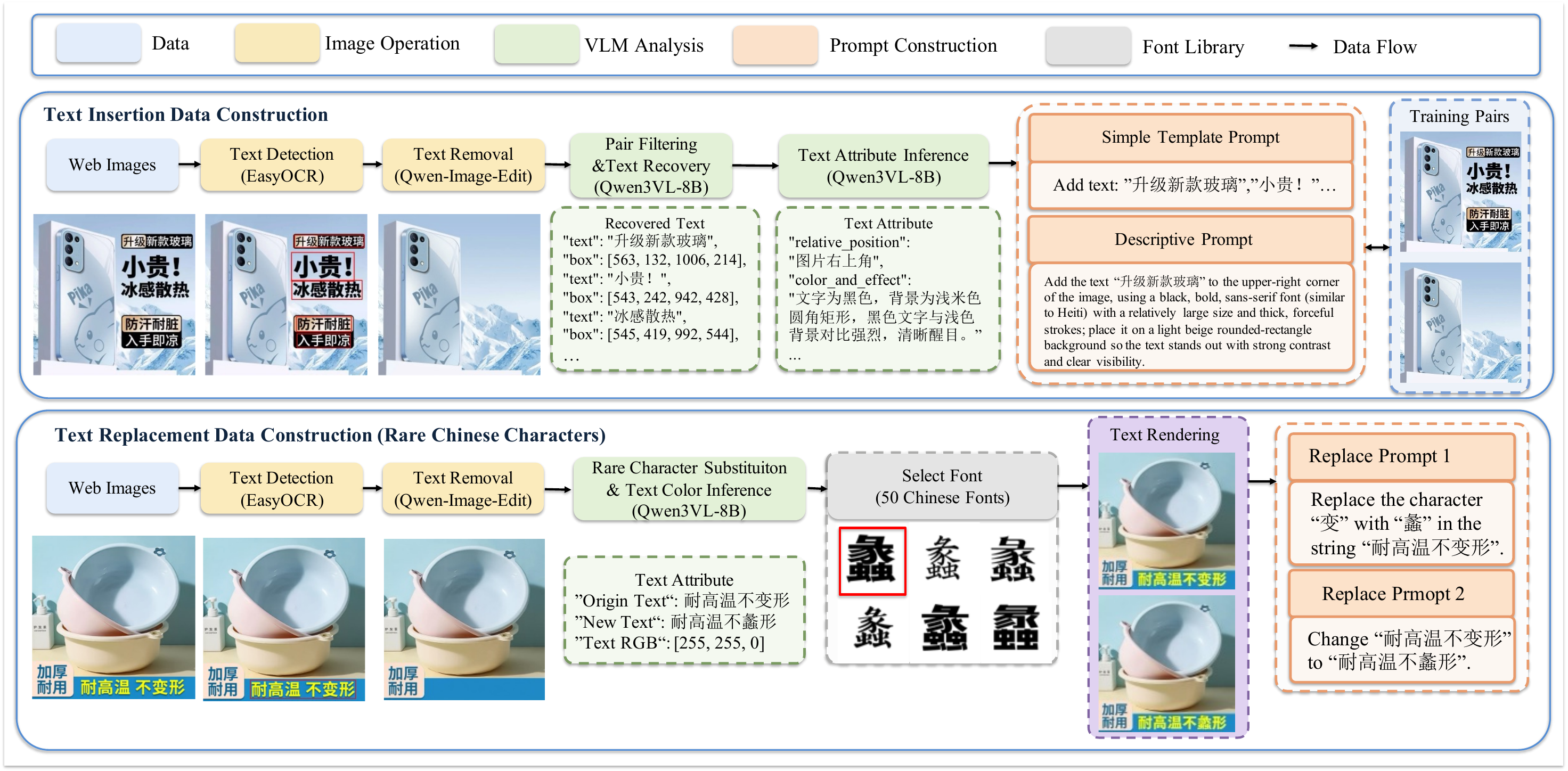}
  \caption{Overview of the data construction pipeline for text insertion and text replacement in OpenTextEdit.}
  \label{fig:data_pipeline}
\end{figure*}

\section{Related Work}
\label{sec:related_work}

\subsection{Visual Text Rendering}

Visual text rendering (VTR), the generation of legible and semantically accurate text within images, has become a key capability of modern generative models. Performance in the generation-from-scratch setting has improved substantially, with recent models achieving high fidelity for dense multilingual text synthesis.

Existing approaches employ two primary strategies. The first injects auxiliary constraints into diffusion models through specialized modules: glyph conditions~\cite{anytext2,glyphcontrol,Glyphdraw2,brushyourtext} for morphological control and layout guidance~\cite{textdiffuser,dreamtext,postermaker,textctrl} for spatial precision. The second improves text encoder designs using character-level tokens or tokenizer-free architectures~\cite{udifftext,textdiffuser2,Glyph-byt5-v2} to preserve fine-grained textual information. Recent large-scale models~\cite{flux,SD3.5,Seeddream4.0,QwenImage} have achieved substantial text-rendering capability without relying on explicit glyph conditions.

However, image editing presents fundamental challenges absent in generation-from-scratch: preserving source content, including foreground subjects, background context, and composition, while inserting or replacing text. Existing editing approaches, including AnyText~\cite{anytext}, AnyText2~\cite{anytext2}, TextCtrl~\cite{textctrl}, GlyphMastero~\cite{glyphmastero}, PosterMaker~\cite{postermaker}, RepText~\cite{Reptext}, and FireRed-Image-Edit~\cite{team2026firered}, typically rely on complex auxiliary inputs such as glyph encoders, prior guidance, and external layout specifications. Despite these efforts, they lack mechanisms for multi-span semantic supervision, fall short in glyph fidelity for complex characters, and do not jointly optimize span-level and character-level quality.

\subsection{Reinforcement Learning for Visual Text Rendering}

Reinforcement learning with scalar reward feedback has emerged as an effective post-training mechanism for improving generative model quality~\cite{unifiedReward}. In the context of VTR, recent works have explored RL-based fine-tuning driven by OCR-derived rewards. Examples include Seedream 2.0~\cite{seedream2.0} and Seedream 3.0~\cite{Seedream3.0}, which incorporate text accuracy rewards during post-training for bilingual rendering; X-Omni~\cite{X-Omni_RL_and_LongText}, which applies RL to discrete autoregressive generation with OCR-based long-text rewards; and BLIP3o-NEXT~\cite{blip3o-next}, which leverages OCR rewards in a generation post-training pipeline.

These methods demonstrate that RL supervision can improve text accuracy in generation-from-scratch settings. However, they share a fundamental limitation that becomes acute in the editing context. Their reward signals rely on standard OCR models~\cite{PPOCRV5,got_ocr2.0} or vision-language models~\cite{Qwen2.5-VL} that operate at the string level, evaluating only character recognition accuracy and lacking fine-grained structural perception. Consequently, such rewards are insensitive to glyph-level defects, including missing strokes, extra components, and local distortions that compromise legibility without failing recognition. Existing RL frameworks for VTR do not distinguish span-level textual quality from glyph-level fidelity, collapsing both into a single reward signal that inadequately supervises text insertion and text replacement.

\begin{figure*}[t]
  \centering
  \includegraphics[width=0.8\textwidth]{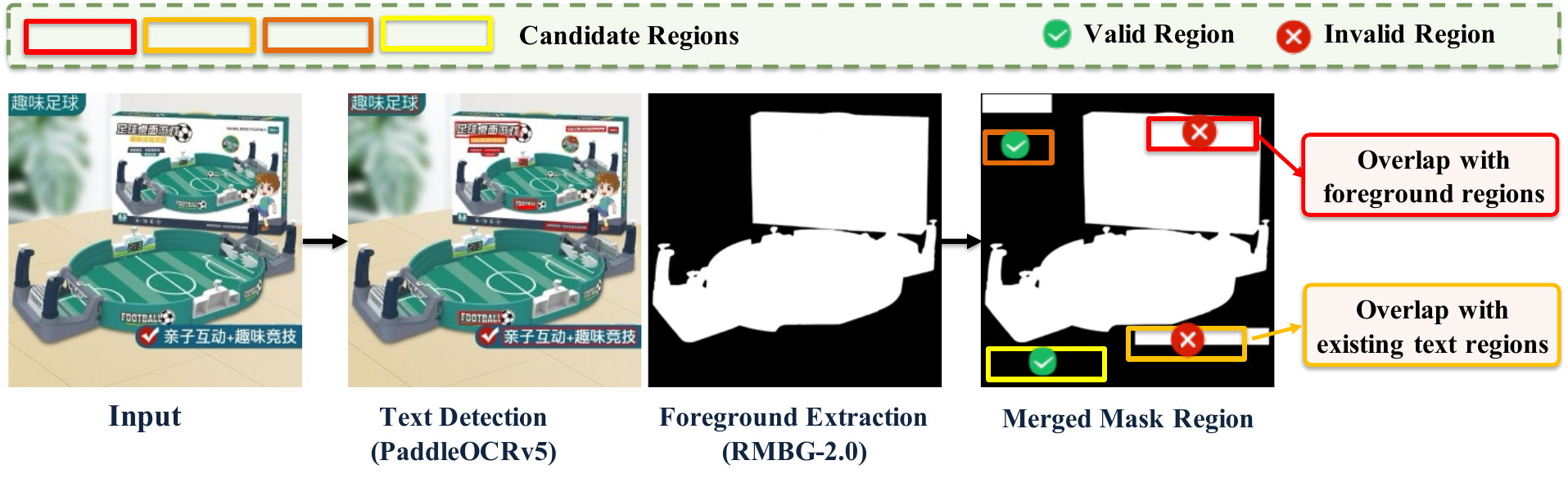}
  \caption{Valid-span filtering for product-poster text insertion. A detected span is retained when its overlap with pre-existing text is below $\tau_{\mathrm{bbox}}$ and its overlap with the salient-product mask $m$ is below $\tau_{\mathrm{mask}}$. This criterion implements the non-overlap preference used in our target setting rather than a universal poster-layout rule.}
  \label{fig:span_filter}
\end{figure*}

\section{Methodology}
\label{sec:methodology}

\subsection{Preliminaries}
\label{sec:prelim}

\subsubsection{DiffusionNFT}
\label{sec:diffusion_nft}

We adopt DiffusionNFT as the online reinforcement learning algorithm for policy optimization. Let $v_\theta(x_t,t,c)$ denote the trainable velocity field and $v^{\mathrm{old}}(x_t,t,c)$ the sampling policy. Given an online sample $x_0$, a task-specific reward model produces a normalized quality score $r \in [0,1]$. DiffusionNFT constructs implicit positive and negative policies as
$v_\theta^{+}=(1-\beta)v^{\mathrm{old}}+\beta v_\theta$ and
$v_\theta^{-}=(1+\beta)v^{\mathrm{old}}-\beta v_\theta$,
where $\beta>0$ controls the update magnitude. The resulting optimization objective is
\begin{equation}
\begin{aligned}
\mathcal{L}_{\mathrm{NFT}}
&= \mathbb{E}_{\substack{(x_0,c)\sim\pi^{\mathrm{old}},\,t;\\
                          x_t\sim q(\cdot\mid x_0,t)}}
\Big[
r\,\lVert v_\theta^{+}(x_t,t,c)-v\rVert_2^2 \\
&\hspace{3.2em}
+(1-r)\,\lVert v_\theta^{-}(x_t,t,c)-v\rVert_2^2
\Big],
\end{aligned}
\label{eq:diffusion_nft}
\end{equation}
where $v$ is the target velocity and $q(x_t \mid x_0,t)$ denotes the forward noising process. The objective assigns greater weight to the positive policy term for high-reward samples and to the negative policy term for low-reward samples. We instantiate $r$ with task-specific signals tailored to text insertion and localized text editing in product posters.

\subsubsection{Task Definition}
\label{sec:task_definition}

We consider two forms of text editing in product posters: \emph{text insertion} and \emph{localized text replacement}. Let $\tau \in \{\mathrm{ins},\mathrm{rep}\}$ denote the task identity. Given a source poster $x^{\mathrm{src}}$, an editing instruction $c$, and a task-specific target $y$, the model generates an edited poster $\hat{x}$ that renders the requested text while preserving product appearance, background content, and non-target layout elements. For insertion ($\tau=\mathrm{ins}$), the target is a set of text spans $y=\{y_i\}_{i=1}^{N}$ to be placed within the available poster layout. For replacement ($\tau=\mathrm{rep}$), the target is a glyph $y^{\star}$ to be rendered in a localized source-text region. We use a simple instruction template and a descriptive attribute-aware template for insertion, together with two operation-specific templates for replacement. These formats expose the model to complementary levels of instruction specificity while retaining an unambiguous editing target. Our goal is to learn a single policy whose reward is selected according to the requirements of each task.

\subsection{Text Editing Data Construction}
\label{sec:data_pipeline}

As illustrated in Figure~\ref{fig:data_pipeline}, we construct task-specific training pairs that reflect the visual and textual constraints of product posters. For insertion, we collect candidate poster images from large-scale web data, localize existing text with OCR, and remove the detected regions using Qwen-Image-Edit-2511. A vision-language model (VLM) filters unreliable edits and retains pairs consisting of an original poster and its text-removed counterpart. By comparing the paired images, the VLM recovers the target spans and annotates visual attributes such as color, position, and font style. We convert these annotations into a simple prompt specifying the insertion target and a descriptive prompt that additionally encodes appearance and layout cues. The resulting pairs supervise both faithful text rendering and compatibility with the existing product-poster composition.

For localized replacement, we construct a complementary subset centered on low-frequency Chinese characters, whose complex structures remain difficult for general-purpose editors. Starting from OCR-localized text regions, we remove a selected region, substitute its content with a target character, and render the character using one of 50 Chinese fonts collected from online sources. This controlled procedure yields paired source and target posters while retaining the surrounding local context. We then generate two replacement-specific prompts that explicitly identify the source and target text, thereby reducing ambiguity in localized editing. We use synthetic rendering only for the replacement subset: insertion additionally requires learning poster-level placement and appearance, whereas localized replacement primarily benefits from controlled supervision of character structure and regional consistency.

\subsection{Text-Span-Level Reward}
\label{sec:text_span_reward}

The insertion reward evaluates whether the edited poster contains the requested spans without introducing spurious text or violating the non-overlap preference adopted in our product-poster setting. Given target spans $y=\{y_i\}_{i=1}^{N}$, we apply OCR to obtain detections $\{(u_j,b_j,s_j)\}_{j=1}^{M}$, where $u_j$, $b_j$, and $s_j$ denote the recognized string, bounding box, and confidence score, respectively. Let $\mathcal{B}^{\mathrm{src}}=\{\bar{b}_k\}$ denote source-poster text boxes, and let $m$ denote the salient-product mask obtained with BiRefNet~\cite{BiRefNet}. We retain detections whose overlap with both pre-existing text and the product region is below predefined thresholds, yielding the valid span set
\begin{equation}
\mathcal{V}=\left\{(v_j,\tilde{b}_j)\,\middle|\,
\begin{aligned}
&\max_{\bar{b}_k\in\mathcal{B}^{\mathrm{src}}}
\frac{\lvert\tilde{b}_j\cap\bar{b}_k\rvert}
     {\min(\lvert\tilde{b}_j\rvert,\lvert\bar{b}_k\rvert)}
<\tau_{\mathrm{bbox}},\\
&\frac{\lvert\tilde{b}_j\cap m\rvert}{\lvert\tilde{b}_j\rvert}
<\tau_{\mathrm{mask}}
\end{aligned}
\right\},
\label{eq:span_filter}
\end{equation}
where $\tau_{\mathrm{bbox}}=0.5$ and $\tau_{\mathrm{mask}}=0.5$ in our implementation. This filtering rule operationalizes a controllable layout preference for the product posters considered here; it is not intended as a universal design principle, since intentional text--product overlap can be appropriate in other poster styles.

For each target span $y_i$, we compute similarity against retained OCR spans using normalized edit distance. Let $\phi(\cdot)$ remove whitespace and normalize case, and define
\begin{equation}
\begin{aligned}
s_{ij}
&=1-\frac{d_{\mathrm{Lev}}\!\left(\phi(y_i),\phi(v_j)\right)}
{\max\!\left\{\lvert\phi(y_i)\rvert,\lvert\phi(v_j)\rvert\right\}},\\
r_i
&=\begin{cases}
1,
& \substack{\exists\,v_j\in\mathcal{V}:\ 
\phi(y_i)\subseteq\phi(v_j)\ \text{or}\\
\phi(v_j)\subseteq\phi(y_i)},\\
\max\limits_{v_j\in\mathcal{V}}s_{ij},
& \max\limits_{v_j\in\mathcal{V}}s_{ij}\ge\tau_{\mathrm{miss}},\\
0, & \text{otherwise}.
\end{cases}
\end{aligned}
\label{eq:span_match}
\end{equation}
where $d_{\mathrm{Lev}}$ is the Levenshtein distance and $\tau_{\mathrm{miss}}=0.5$. We then combine semantic fidelity and coverage as
\begin{equation}
R_{\mathrm{span}}=\underbrace{\frac{1}{N}\sum_{i=1}^{N} r_i}_{R_{\mathrm{sim}}}\cdot
\underbrace{\max\left(0,1-\frac{U_{\mathrm{tar}}+U_{\mathrm{ocr}}}{\max(N,|\mathcal{V}|)}\right)}_{R_{\mathrm{cov}}},
\label{eq:span_reward}
\end{equation}
where $U_{\mathrm{tar}}$ and $U_{\mathrm{ocr}}$ denote the numbers of unmatched target spans and unmatched OCR detections, respectively. The similarity term measures target-span fidelity, while the coverage term penalizes omissions and unintended text generation. Together with the filtering in Equation~\ref{eq:span_filter}, this reward assesses content coverage and collision avoidance; it does not attempt to model the full aesthetics of poster layout.

To reduce OCR-oriented reward exploitation and preserve the source poster, we introduce a gated structural regularizer that becomes active only after span-level accuracy exceeds a prescribed threshold:
\begin{equation}
\begin{aligned}
R_{\mathrm{text}}
&=R_{\mathrm{span}}
+\mathbf{1}\!\left[R_{\mathrm{span}}>\tau_{\mathrm{ssim}}\right]\\
&\quad\cdot\operatorname{SSIM}\!\left(
\psi(\hat{x},\mathcal{V}),
\psi(x^{\mathrm{src}},\mathcal{V})
\right),
\end{aligned}
\label{eq:text_reward_final}
\end{equation}
where $\psi(\cdot,\mathcal{V})$ masks the retained text boxes in $\mathcal{V}$ by whitening them before comparison, and $\tau_{\mathrm{ssim}}=0.5$. Consequently, the optimization first prioritizes rendering the requested text; once this condition is met, the SSIM term penalizes unintended changes to products, backgrounds, and other non-text poster content.

\subsection{Glyph-Level Reward}
\label{sec:glyph_reward}

Although $R_{\mathrm{text}}$ measures span-level correctness, it is comparatively insensitive to fine structural defects within an individual character. We therefore define a glyph-level reward for localized replacement samples ($\tau=\mathrm{rep}$) using graded recognition evidence from the edited character region. Let $y^{\star}$ denote the target glyph and $b^{\star}=[x_1,y_1,x_2,y_2]$ its annotated bounding box. We first crop the edited poster $\hat{x}$ to this region,
\begin{equation}
\hat{x}_{\mathrm{char}} = \operatorname{Crop}(\hat{x}, b^{\star}),
\label{eq:glyph_crop}
\end{equation}
and feed the crop into a PaddleOCR-v5 recognizer. Instead of relying only on the final decoded string, we use the full CTC posterior matrix
\begin{equation}
P \in [0,1]^{T \times |\mathcal{C}|},
\end{equation}
where $T$ is the temporal length of the recognition sequence, $\mathcal{C}$ is the OCR character vocabulary, and $P_{t,k}$ denotes the softmax probability assigned to character $k \in \mathcal{C}$ at time step $t$.

\begin{figure}[t]
  \centering
  \includegraphics[width=1.0\linewidth]{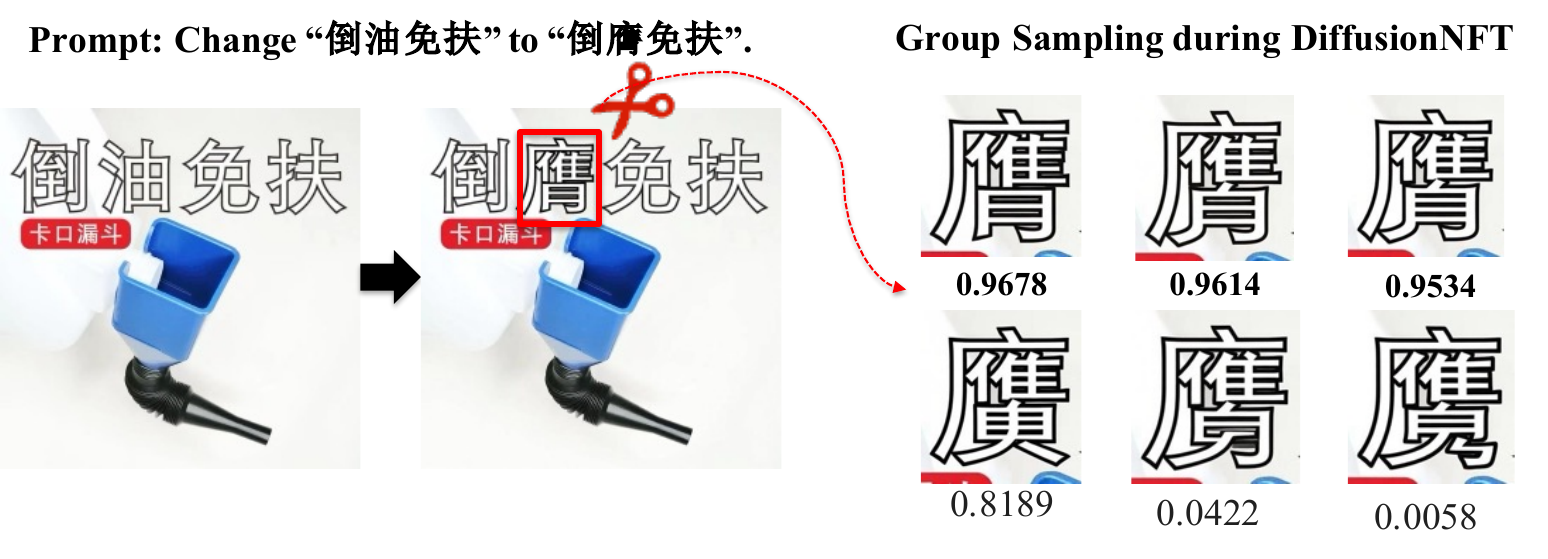}
  \caption{Comparison of binary OCR correctness and the proposed graded glyph signal. Whereas a binary reward retains only the final correctness decision, the target-glyph CTC posterior preserves continuous recognition evidence and provides a more informative signal for optimizing glyph fidelity.}
  \label{fig:glyph_score_signal}
\end{figure}

As illustrated in Figure~\ref{fig:glyph_score_signal}, the posterior retains graded evidence that is discarded by binary OCR correctness. Let $\operatorname{id}(y^{\star})$ be the vocabulary index of $y^{\star}$. We define the glyph reward as the maximum posterior probability assigned to the target glyph across all CTC time steps:
\begin{equation}
R_{\mathrm{glyph}} = \max_{1 \le t \le T} P_{t,\operatorname{id}(y^{\star})}.
\label{eq:glyph_reward}
\end{equation}
If the recognizer produces multiple candidate sequences, we concatenate them along the temporal dimension before applying the maximum. The resulting score increases with the recognizer's support for the target glyph and provides a continuous optimization signal under missing strokes, structural deformation, or confusion with visually similar characters. This reward is particularly suited to localized editing of low-frequency Chinese characters, for which exact-match supervision is sparse.

\subsection{Task-Specific Reward Assignment}
\label{sec:unified_reward}

The two rewards address different editing operations and are therefore not summed for an individual sample. We train TextRefine on a mixed stream of insertion and localized-replacement examples and select the scalar DiffusionNFT reward according to the task identity:
\begin{equation}
R(\hat{x},\tau,y,b^{\star})=
\begin{cases}
R_{\mathrm{text}}, & \tau = \mathrm{ins},\\
R_{\mathrm{glyph}}, & \tau = \mathrm{rep}.
\end{cases}
\label{eq:task_specific_reward}
\end{equation}
Thus, insertion samples receive span- and preservation-aware supervision through $R_{\mathrm{text}}$, whereas replacement samples receive character-structure supervision through $R_{\mathrm{glyph}}$. This task-conditioned assignment allows a single editing policy to learn from both operations without conflating their distinct quality criteria.

\section{Experiments}
\label{sec:experiments}

\subsection{Implementation Details}

\paragraph{Data and Supervised Fine-Tuning.}
We evaluate TextRefine under a common protocol for insertion and localized replacement in product posters, with all training and evaluation images resized to $1024\times1024$. OpenTextEdit contains 50K insertion images and 50K localized-replacement images. Pairing each image with two instruction variants yields 200K prompt--image instances for supervised fine-tuning. We initialize from Qwen-Image-Edit-2511 and optimize all model parameters using DeepSpeed ZeRO-3 with bf16 mixed precision. Training uses a global batch size of 8 and a learning rate of $1\times10^{-5}$ for approximately 20 epochs on 48 NVIDIA A800 GPUs.

\paragraph{Hardness-Aware RL Data Selection.}
To improve the informativeness of online optimization, we prioritize challenging examples that remain feasible for the SFT policy. Specifically, we consider insertion examples containing more than three target spans and replacement examples whose target characters contain more than 20 strokes. For each candidate, the SFT model generates five rollouts using 15 denoising steps. We compute the corresponding task-specific rewards and retain, for each operation, the 1,000 examples with the largest reward variance. This strategy emphasizes semi-hard cases for which the policy is capable of producing valid edits but remains unstable across rollouts.

\paragraph{Reinforcement Learning.}
Starting from the SFT checkpoint, we apply DiffusionNFT with LoRA rank 64 and scaling factor 128. Each update uses groups of 16 samples, with 48 distinct prompts sampled per epoch. Rollouts use 15 denoising steps, and optimization is performed on four sampled timesteps from each trajectory. We train for approximately two epochs on 16 NVIDIA A800 GPUs using a learning rate of $3.0\times10^{-4}$.

\begin{figure*}[t]
  \centering
  \includegraphics[width=0.8\textwidth]{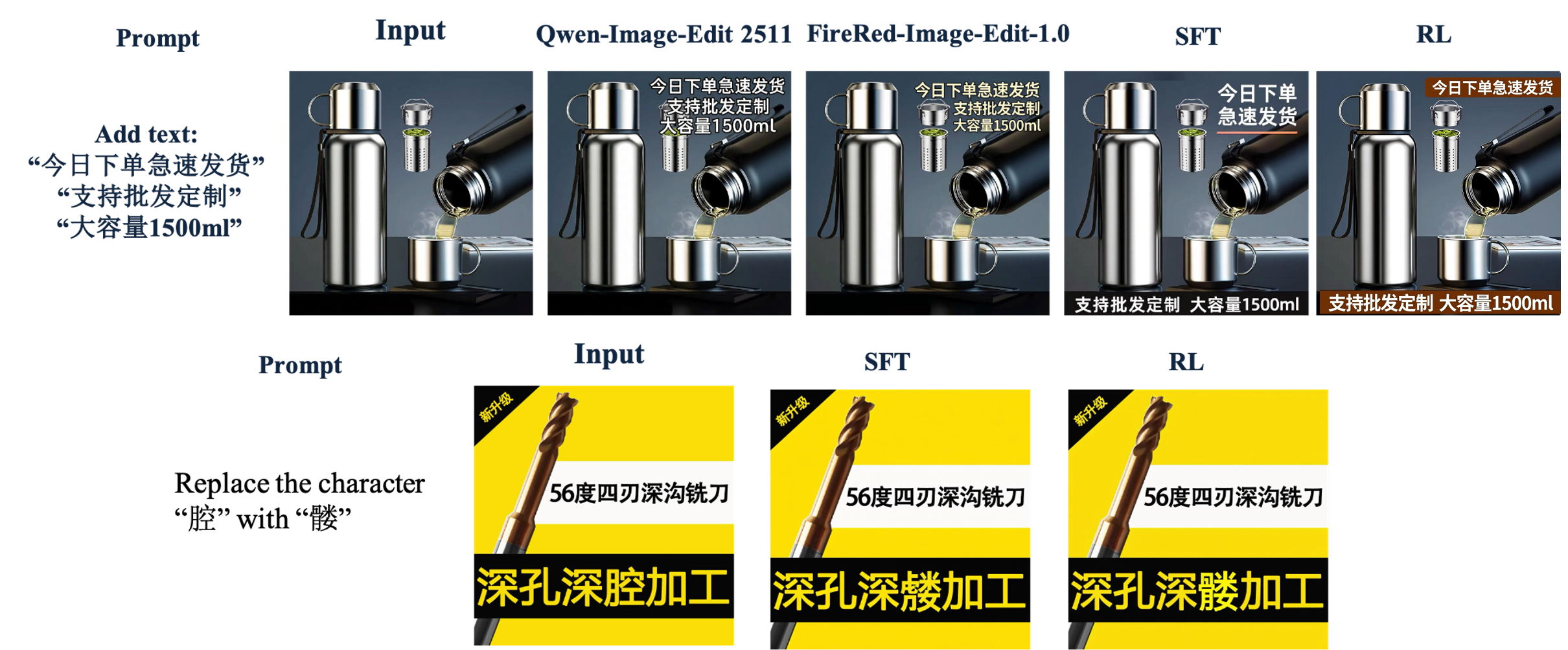}
  \caption{Qualitative comparison of TextRefine with the evaluated baselines on product-poster text insertion (top) and localized replacement (bottom). TextRefine more faithfully renders the requested content while respecting the target poster layout and character structure; baseline outputs exhibit product occlusion, unintended text, or glyph distortion.}
  \label{fig:qualitative}
\end{figure*}

\subsection{Evaluation Metrics}

We construct task-specific evaluation subsets from OpenTextEdit. The insertion benchmark contains 629 examples with simple prompts and 1,235 examples with descriptive prompts, while the localized-replacement benchmark contains 200 examples evaluated with a fixed replacement prompt. We report operation-specific text metrics together with FID, SSIM, and PSNR to characterize distributional quality and preservation of source-poster content. Because the benchmarks follow the target product-poster distribution, the results measure in-domain performance and should not be interpreted as evidence of unrestricted cross-domain or multilingual generalization.

\paragraph{Text Insertion Benchmark.}
Each example contains a source poster, a target span set $Y_j=\{y_{j,i}\}_{i=1}^{N_j}$, and an edited output. We apply PaddleOCR-v5 to the source and edited posters and remove detections associated with pre-existing source text using bounding-box overlap, leaving candidate inserted spans. Target spans are parsed from the instruction, and a VLM aligns them with OCR detections while accounting for token fragmentation and minor recognition deviations (similarity $>0.7$). For example $j$, let $M_j$, $P_j$, and $U_j$ denote the numbers of \emph{matched}, \emph{partial}, and \emph{missed} targets, respectively, and let $E_j$ denote unmatched extra detections; thus $N_j=M_j+P_j+U_j$.

The sample-level insertion score is
\begin{equation}
s_j^{\mathrm{ins}} = \max\!\left(0,\, \frac{M_j + \alpha P_j - E_j}{N_j}\right),
\end{equation}
where $\alpha\in[0,1]$ weights partial matches. Over a benchmark with $B$ samples, we report
\begin{equation}
\mathrm{Score}_{\mathrm{ins}} = \frac{1}{B}\sum_{j=1}^{B} s_j^{\mathrm{ins}}.
\end{equation}

\paragraph{Localized Text Replacement Benchmark.}
Each example consists of a source poster, a target character $y_j^{\star}$, and an annotated character box $b_j^{\star}$. We crop the edited poster around $b_j^{\star}$ with a small margin and apply PaddleOCR-v5 to the localized region. Let $t_j$ denote the concatenated OCR output. We define the correctness indicator as
\begin{equation}
z_j = \mathbb{I}\!\left[y_j^{\star} \in t_j\right],
\end{equation}
where $\mathbb{I}[\cdot]$ is the Iverson bracket that equals 1 if the target character is present in the OCR output and 0 otherwise. The replacement accuracy is then
\begin{equation}
\mathrm{Acc}_{\mathrm{rep}} = \frac{1}{B}\sum_{j=1}^{B} z_j,
\end{equation}
which measures the proportion of examples for which the target character is recognized within the localized replacement region. This protocol focuses on single-character editing and does not evaluate multi-character rewriting.

\subsection{Main Results}

\subsubsection{Comparison with Baselines}

We compare TextRefine with four representative baselines. Qwen-Image-Edit-2511~\cite{QwenImage} is the general-purpose instruction-based editor used for initialization, while FireRed-Image-Edit-1.0~\cite{team2026firered} provides a strong visual-text editing baseline. We additionally evaluate AnyText2~\cite{anytext2}, which supports attribute-controllable visual text generation and editing, and PosterMaker~\cite{postermaker}, which is specifically designed for accurate text rendering in product posters. Together, these methods cover general instruction-based editing, specialized visual-text editing, and product-poster-oriented generation. Table~\ref{tab:insertion_results} reports the main comparison under both prompt settings, and Figure~\ref{fig:qualitative} provides qualitative comparisons. Our claims are restricted to the evaluated baselines and the OpenTextEdit product-poster benchmarks.

As shown in Table~\ref{tab:insertion_results}, TextRefine (RL) obtains the strongest overall results among the evaluated methods under both prompt settings. Relative to its SFT initialization, reward optimization increases the matched-span rate and insertion score while reducing partial matches, missed spans, and extra OCR detections. The accompanying gains in SSIM and PSNR indicate improved preservation of non-target poster content. These results support the role of the span reward in suppressing omissions and unintended text, and of the gated structural term in limiting changes outside the inserted regions.

\begin{table*}[t]
\centering
\small
\caption{Quantitative comparison on text insertion benchmark under simple template prompt (top, 629 samples) and descriptive prompt (bottom, 1235 samples). $\uparrow$: higher is better; $\downarrow$: lower is better. Best results in \textbf{bold}.}
\label{tab:insertion_results}
\begin{tabular}{lcccccccc}
\toprule
Method & Match (\%)$\uparrow$ & Partial (\%)$\downarrow$ & Miss (\%)$\downarrow$ & Extra$\downarrow$ & Score$\uparrow$ & FID$\downarrow$ & SSIM$\uparrow$ & PSNR$\uparrow$ \\
\midrule
\multicolumn{9}{c}{\textit{Simple Template Prompt (629 samples)}} \\
\midrule
Qwen-Image-Edit-2511 & 76.8 & 8.3 & 14.9 & 26 & 0.7829 & 36.7330 & 0.8016 & 14.8536 \\
FireRed-Image-Edit-1.0 & 82.2 & 6.2 & 11.6 & 32 & 0.8277 & 30.5686 & 0.7825 & 14.9120 \\
TextRefine (SFT) & 87.1 & 7.2 & 5.7 & 52 & 0.8526 & 24.3464 & 0.8232 & 15.9917 \\
\textbf{TextRefine (RL)} & \textbf{91.9} & \textbf{4.3} & \textbf{3.8} & \textbf{18} & \textbf{0.9174} & \textbf{24.2191} & \textbf{0.8303} & \textbf{16.0774} \\
\midrule
\multicolumn{9}{c}{\textit{Descriptive Prompt (1235 samples)}} \\
\midrule
Qwen-Image-Edit-2511 & 79.4 & 7.9 & 12.6 & 102 & 0.7893 & 48.9229 & 0.6412 & 12.3042 \\
FireRed-Image-Edit-1.0 & 82.5 & 7.2 & 10.3 & 76 & 0.8216 & 36.2180 & 0.6782 & 13.3152 \\
TextRefine (SFT) & 87.8 & 2.8 & 9.4 & 82 & 0.8419 & 32.5287 & 0.7364 & 14.7352 \\
\textbf{TextRefine (RL)} & \textbf{90.0} & \textbf{1.2} & \textbf{8.7} & \textbf{63} & \textbf{0.8973} & \textbf{31.9462} & \textbf{0.7428} & \textbf{15.1628} \\
\bottomrule
\end{tabular}

\vspace{0.8em}
\setcounter{table}{2}
\caption{Ablation on text insertion reward components (simple template prompt, 629 samples). $\uparrow$: higher is better; $\downarrow$: lower is better. Best results in \textbf{bold}.}
\label{tab:ablation_insertion}
\resizebox{0.96\textwidth}{!}{%
\begin{tabular}{lcccccccc}
\toprule
Method & Match (\%)$\uparrow$ & Partial (\%)$\downarrow$ & Miss (\%)$\downarrow$ & Extra$\downarrow$ & Score$\uparrow$ & FID$\downarrow$ & SSIM$\uparrow$ & PSNR$\uparrow$ \\
\midrule
SFT (No RL)                          & 87.1 & 7.2 & 5.7 & 52 & 0.8526 & 24.3464 & 0.8232 & 15.9917 \\
$R_{\mathrm{span}}$                  & 88.6 & 4.8 & 6.7 & 23 & 0.8800 & 24.2653 & 0.8357 & 16.0874 \\
\textbf{$R_{\mathrm{text}}$ (with gated SSIM)} & \textbf{91.9} & \textbf{4.3} & \textbf{3.8} & \textbf{18} & \textbf{0.9174} & \textbf{24.1872} & \textbf{0.8427} & \textbf{16.1573} \\
\bottomrule
\end{tabular}%
}
\setcounter{table}{1}
\end{table*}

\paragraph{Localized Text Replacement Results.}
Table~\ref{tab:replacement_results} summarizes performance on localized editing of low-frequency Chinese characters. TextRefine (SFT) already improves substantially over the evaluated baselines, showing the benefit of task-specific replacement data for structurally complex glyphs. Applying the glyph reward yields a further improvement, supporting the use of the target-character CTC posterior as an informative optimization signal for this benchmark. Detailed reward ablations for both editing operations are presented below.

\begin{table}[!htbp]
\centering
\scriptsize
\captionsetup{skip=2pt}
\caption{Accuracy on the localized rare-character editing benchmark (200 samples). Higher is better. Best result in \textbf{bold}.}
\label{tab:replacement_results}
\resizebox{0.60\columnwidth}{!}{%
\begin{tabular}{lc}
\toprule
Method & Acc (\%)$\uparrow$ \\
\midrule
Qwen-Image-Edit-2511        & 48.0 \\
FireRed-Image-Edit-1.0      & 55.5 \\
TextRefine (SFT)            & 69.5 \\
\textbf{TextRefine (RL)}    & \textbf{74.5} \\
\bottomrule
\end{tabular}%
}
\end{table}

\subsubsection{Ablation Studies}
\paragraph{Text Insertion.}
Table~\ref{tab:ablation_insertion} shows that optimizing $R_{\mathrm{span}}$ reduces extra OCR detections and improves the insertion score relative to SFT, indicating that the similarity and coverage terms discourage missing and unintended text. Adding the gated SSIM term to form $R_{\mathrm{text}}$ further improves matched-span accuracy and source-poster similarity. Because the structural term is activated only after $R_{\mathrm{span}}$ exceeds the threshold, it complements rather than replaces the text-centric objective.

\paragraph{Localized Text Replacement.}
Table~\ref{tab:ablation_replacement} shows that $R_{\mathrm{glyph}}$ improves accuracy over both SFT and the binary OCR reward. This result indicates that retaining the target-character posterior supplies useful graded supervision for the structurally complex, low-frequency characters represented in the benchmark.

\begin{table}[!htbp]
\centering
\scriptsize
\captionsetup{skip=2pt}
\setcounter{table}{3}
\caption{Ablation on localized text replacement rewards (rare-character benchmark, 200 samples). Higher is better. Best results in \textbf{bold}.}
\label{tab:ablation_replacement}
\resizebox{0.60\columnwidth}{!}{%
\begin{tabular}{lc}
\toprule
Method & Acc (\%)$\uparrow$ \\
\midrule
SFT (No RL)                          & 69.5 \\
Binary OCR Reward                    & 73.0 \\
\textbf{$R_{\mathrm{glyph}}$ (Ours)} & \textbf{76.0} \\
\bottomrule
\end{tabular}%
}
\end{table}

\section{Conclusion}
\label{sec:conclusion}

We presented \textbf{TextRefine}, a task-aligned post-training framework for text editing in product posters. Its span-level insertion reward jointly promotes target-text fidelity, coverage, reliable placement, and preservation of non-target content, while its CTC-posterior glyph reward provides graded structural supervision for localized character replacement. We also introduced OpenTextEdit, a 100K-image dataset with multi-text poster layouts, detailed text attributes, product masks, and challenging low-frequency Chinese characters. On the evaluated benchmarks, TextRefine improves insertion accuracy, source-poster preservation, and localized replacement accuracy over its SFT initialization and the compared baselines. Evaluation remains limited to OpenTextEdit, Chinese single-character replacement, and PaddleOCR-v5; future work will address external, multilingual, and multi-character settings.

\bibliography{cofirender_references}

\clearpage

\section{Appendix}
\setcounter{table}{0}
\setcounter{figure}{0}

\subsection{A.1 More Dataset and Benchmark Details}
\label{app:dataset-benchmark-details}

\subsubsection{Product Category Distribution}
\label{app:product-category-distribution}

We group products into five functional domains: \textit{Wearable and Fashion Goods} (e.g., jewelry, footwear, and handbags), \textit{Home and Lifestyle Products} (e.g., bedding, drinkware, and cookware), \textit{Food and Agricultural Commodities} (e.g., packaged snacks, fresh produce, and seafood), \textit{Beauty and Personal Care Products} (e.g., cosmetics, skincare products, and manicure products), and \textit{Digital and General Consumer Goods} (e.g., phone accessories, toys, and pet supplies). Tables~\ref{tab:product-category-distribution} and~\ref{tab:evaluation-category-distribution} report the corresponding distributions of the approximately 100K-example training set and the 200-example evaluation benchmark, respectively. Representative evaluation examples from the five product domains are shown in Figure~\ref{fig:evaluation-domain-examples}.

\begin{table}[h]
\centering
\caption{Product-category distribution of the approximately 100K-example training set.}
\label{tab:product-category-distribution}
\begin{tabular}{lc}
\toprule
Category & Share (\%) \\
\midrule
Wearable and Fashion Goods            & 24.6 \\
Home and Lifestyle Products           & 21.8 \\
Food and Agricultural Commodities     & 19.3 \\
Beauty and Personal Care Products     & 17.9 \\
Digital and General Consumer Goods    & 16.4 \\
\bottomrule
\end{tabular}
\end{table}

\begin{table}[h]
\centering
\caption{Product-category distribution of the 200-example evaluation benchmark.}
\label{tab:evaluation-category-distribution}
\begin{tabular}{lc}
\toprule
Category & Share (\%) \\
\midrule
Wearable and Fashion Goods            & 24.5 \\
Home and Lifestyle Products           & 22.0 \\
Food and Agricultural Commodities     & 19.5 \\
Beauty and Personal Care Products     & 18.0 \\
Digital and General Consumer Goods    & 16.0 \\
\bottomrule
\end{tabular}
\end{table}

\subsubsection{SFT Training Data Composition}
\label{app:sft-data-composition}

Our SFT corpus contains five tasks spanning text insertion, replacement, and removal: text insertion with simple template prompts (\textbf{TI-S}), text insertion with descriptive prompts (\textbf{TI-D}), general text replacement (\textbf{TR-G}), rare-character replacement (\textbf{TR-R}), and text removal (\textbf{RM}). The rare-text insertion examples are divided approximately equally between TI-S and TI-D. TR-G is generated by directly editing characters with Qwen-Image-Edit-2511 and retaining outputs that pass quality filtering, whereas TR-R uses synthetically rendered low-frequency characters. RM consists of intermediate text-erased images produced during insertion-data construction. Table~\ref{tab:sft-task-distribution} summarizes the resulting composition, and representative training examples for the five tasks are shown in Figure~\ref{fig:sft-task-examples}.

\newpage
\begin{table}[h]
\centering
\small
\setlength{\tabcolsep}{5pt}
\caption{Task composition of the SFT training corpus.}
\label{tab:sft-task-distribution}
\begin{tabular}{lcc}
\toprule
Task & Samples & Share (\%) \\
\midrule
TI-S                                  & 45,829 & 27.90 \\
TI-D                                  & 30,408 & 18.51 \\
TR-G                                  & 10,000 & 6.09 \\
TR-R                                  & 61,148 & 37.23 \\
RM                                    & 16,862 & 10.27 \\
\midrule
\textbf{Total}                       & \textbf{164,247} & \textbf{100.00} \\
\bottomrule
\end{tabular}
\end{table}

\subsubsection{Prompt Templates for Data Construction}
\label{app:prompt-templates}

The OpenTextEdit construction pipeline uses three VLM prompts to recover newly inserted text from paired images, describe the visual attributes of localized text regions, and select rendering colors with sufficient contrast after text removal. We provide the English templates below for reproducibility.

\setcounter{figure}{2}
\begin{appendixpromptbox}{Prompt for Pair Filtering and Text Recovery}
You are an expert in visual text understanding and OCR.

Task Description:
The input consists of two images:

1. Image 1: the edited image, which may contain newly added text to be verified.
2. Image 2: the corresponding pre-edit image, or the original background image with the target text removed.

Your task is to compare the two images and detect only the text content that appears in Image 1 but is absent from Image 2. Text that is present in both images, including background text or unchanged scene text, must be ignored.

Output Requirements:

1. The output must be a strictly valid JSON array.
2. Each element must be a dictionary containing exactly two fields:
   * "text": the recognized newly added text string.
   * "bbox": the pixel-level bounding box in Image 1, formatted as [x1, y1, x2, y2], where [x1, y1] and [x2, y2] are the top-left and bottom-right corners.
3. If the newly added text contains multiple lines, represent each line as a separate dictionary element. The "text" field must not contain newline characters.
4. Do not include explanations, comments, notes, or Markdown code-block markers.
5. The output must be syntactically valid JSON, with no trailing commas, missing brackets, or unclosed dictionaries.

Example Output:
[
  {
    "text": "First line of example text",
    "bbox": [10, 20, 100, 50]
  },
  {
    "text": "Second line of example text",
    "bbox": [10, 60, 100, 90]
  }
]
\end{appendixpromptbox}
\refstepcounter{figure}\label{fig:prompt-pair-filtering}
\noindent\small Figure~\thefigure: \textbf{Prompt for pair filtering and text recovery.} The template identifies newly added text spans and their pixel-level bounding boxes by comparing paired images.\par
\vspace{4pt}

\newpage
\begin{appendixpromptbox}{Prompt for Text Attribute Description}
You are an expert in image analysis, visual typography, and layout understanding.

Task Description:
The input consists of an original image and a list of text regions. Each region is specified by its textual content (`text`) and absolute pixel-level bounding box (`bbox`). Analyze each region from its actual visual appearance and return one dictionary per region in a strictly valid JSON array.

Each dictionary must contain the following fields in exactly this order:

* "text": Copy the input text exactly without modification.
* "bbox": Copy the input bounding box exactly in the format [x1, y1, x2, y2].
* "relative_position": Concisely describe the text position, such as "top-left corner of the image", "above the person", or "middle-right side of the product".
* "color_and_effect": Describe the text color, contrast with the background, and visible effects such as gradient filling, metallic texture, outline, glow, highlight, embossing, or three-dimensional appearance.
* "font_style": Describe typographic properties, including stroke weight, serif or sans-serif style, character type, estimated size, shadow, outline, perspective, and overall aesthetic. When possible, infer a likely font category such as Songti, Heiti, Kaiti, semi-cursive, rounded, or display style.
* "direction_and_tilt": Describe orientation and tilt, including horizontal or vertical layout, estimated rotation, or a curved path.
* "background_features": Describe any panel, color block, gradient, decorative texture, or geometric element behind the text. Use an empty string if no such feature is visible.

Rules:

1. The "text" and "bbox" fields must be identical to the input; do not correct, translate, normalize, or modify them.
2. All visual attributes must be grounded in observable image evidence. Do not hallucinate properties.
3. Return a strictly valid JSON array with exactly one dictionary per input region.
4. Do not include explanations, comments, notes, or Markdown code-block markers.
5. Do not use trailing commas or malformed JSON.

Input Format Example:
[
  {"text": "VANOW", "bbox": [20, 30, 450, 70]},
  {"text": "Large Capacity", "bbox": [50, 300, 200, 400]}
]

Output Format Example:
[
  {
    "text": "VANOW",
    "bbox": [20, 30, 450, 70],
    "relative_position": "Top-left area of the image.",
    "color_and_effect": "Bright gold text with strong contrast against a red-to-purple gradient background.",
    "font_style": "Large bold sans-serif English display font with thick, rigid strokes.",
    "direction_and_tilt": "Horizontally arranged with no visible tilt.",
    "background_features": "A right-tilted parallelogram panel with a red-to-purple gradient."
  }
]
\end{appendixpromptbox}
\refstepcounter{figure}\label{fig:prompt-text-attribute}
\noindent\small Figure~\thefigure: \textbf{Prompt for text attribute description.} The template describes the position, color effects, typography, orientation, and background features of each localized text region.\par
\vspace{4pt}

\newpage
\begin{appendixpromptbox}{Prompt for Text Color Inference}
You are an expert in image color analysis.

Task Description:
The input consists of an image and information for one text region:

* `bbox`: the absolute pixel-level bounding box [x1, y1, x2, y2].
* `font_color_rgb`: the original font color [R, G, B].
* `background_rgb`: the original background-panel color [R, G, B], or null if no panel exists.

The text in the specified region has already been removed, so the current `bbox` contains only the underlying background.

Your task is to:

1. Inspect the current visual color inside `bbox`.
2. Determine whether `font_color_rgb` provides sufficient perceptual contrast with the current region background.
3. Select `render_color_rgb` using the following priority:
   * If `font_color_rgb` has sufficient contrast, use it directly.
   * Otherwise, if `background_rgb` is not null and has sufficient contrast with the current background, use `background_rgb`.
   * If neither color provides sufficient contrast, automatically choose a strongly contrasting color, such as a dark color for a light background or a light color for a dark background.
4. Return `region_bg_rgb`, the dominant current background color inside `bbox`.

Output Requirements:
Return a strictly valid JSON object containing exactly:

* "region_bg_rgb": the dominant current region color [R, G, B].
* "render_color_rgb": the selected rendering color [R, G, B].
* "render_color_source": exactly one of "font_color_rgb", "background_rgb", or "auto_contrast".
* "reason": a brief one-sentence explanation.

Rules:

1. Every RGB value must be an integer array of length three, with components in [0, 255].
2. Output only the JSON object, without explanations, comments, notes, or Markdown code-block markers.
3. Do not use trailing commas or malformed JSON.

Input Format Example:
{
  "bbox": [63, 204, 240, 253],
  "font_color_rgb": [255, 255, 255],
  "background_rgb": [220, 30, 30]
}

Output Format Example:
{
  "region_bg_rgb": [210, 25, 25],
  "render_color_rgb": [255, 255, 255],
  "render_color_source": "font_color_rgb",
  "reason": "The dark red region provides clear contrast with the original white font color."
}
\end{appendixpromptbox}
\refstepcounter{figure}\label{fig:prompt-color-inference}
\noindent\small Figure~\thefigure: \textbf{Prompt for text color inference.} The template selects a rendering color by comparing the original font color, the historical background-panel color, and the current erased-region background.\par

\clearpage
\setcounter{figure}{0}

\begin{figure*}[p]
  \centering
  \includegraphics[width=0.78\textwidth]{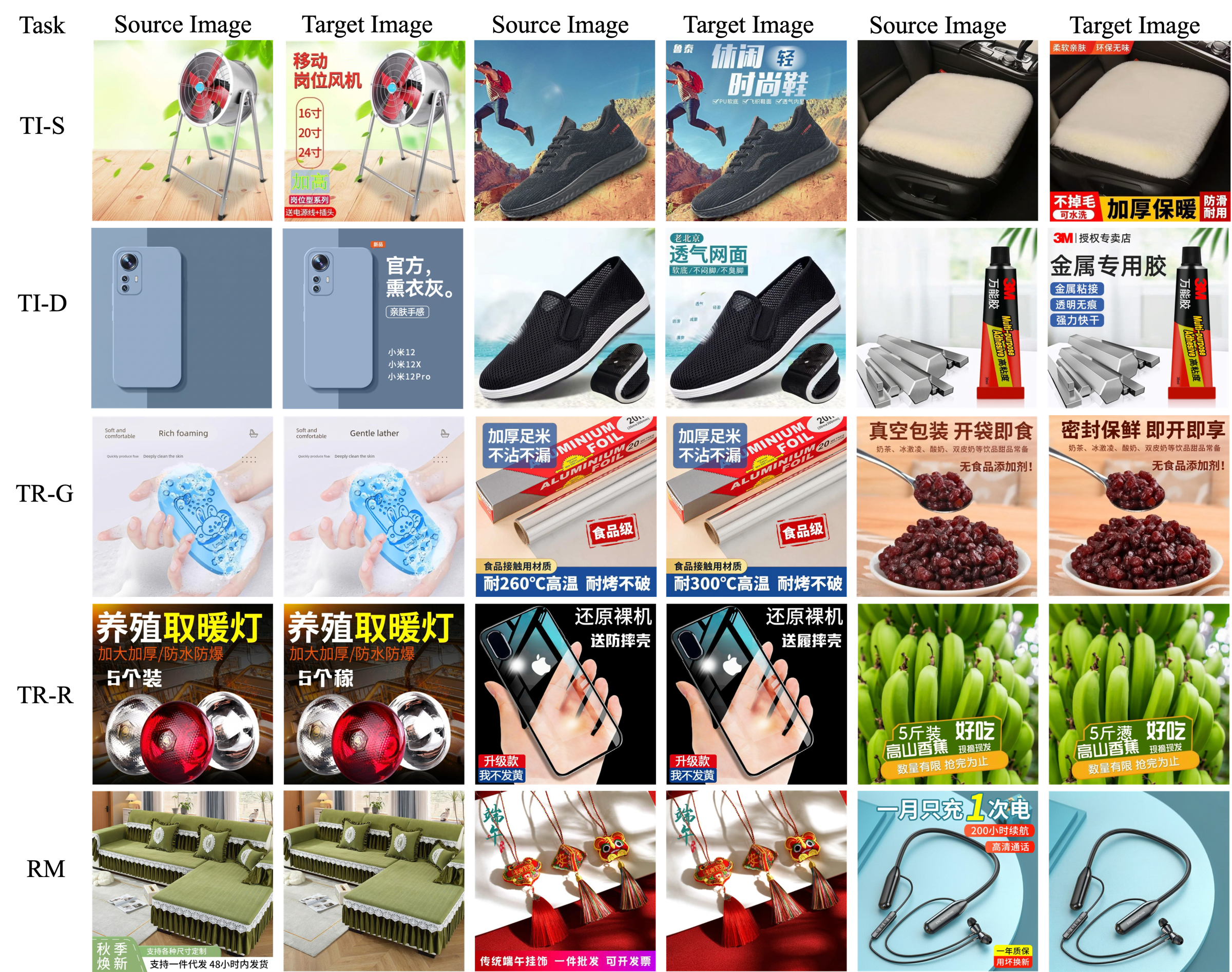}
  \caption{Representative training examples for the five SFT tasks.}
  \label{fig:sft-task-examples}
\end{figure*}

\begin{figure*}[p]
  \centering
  \includegraphics[width=0.78\textwidth]{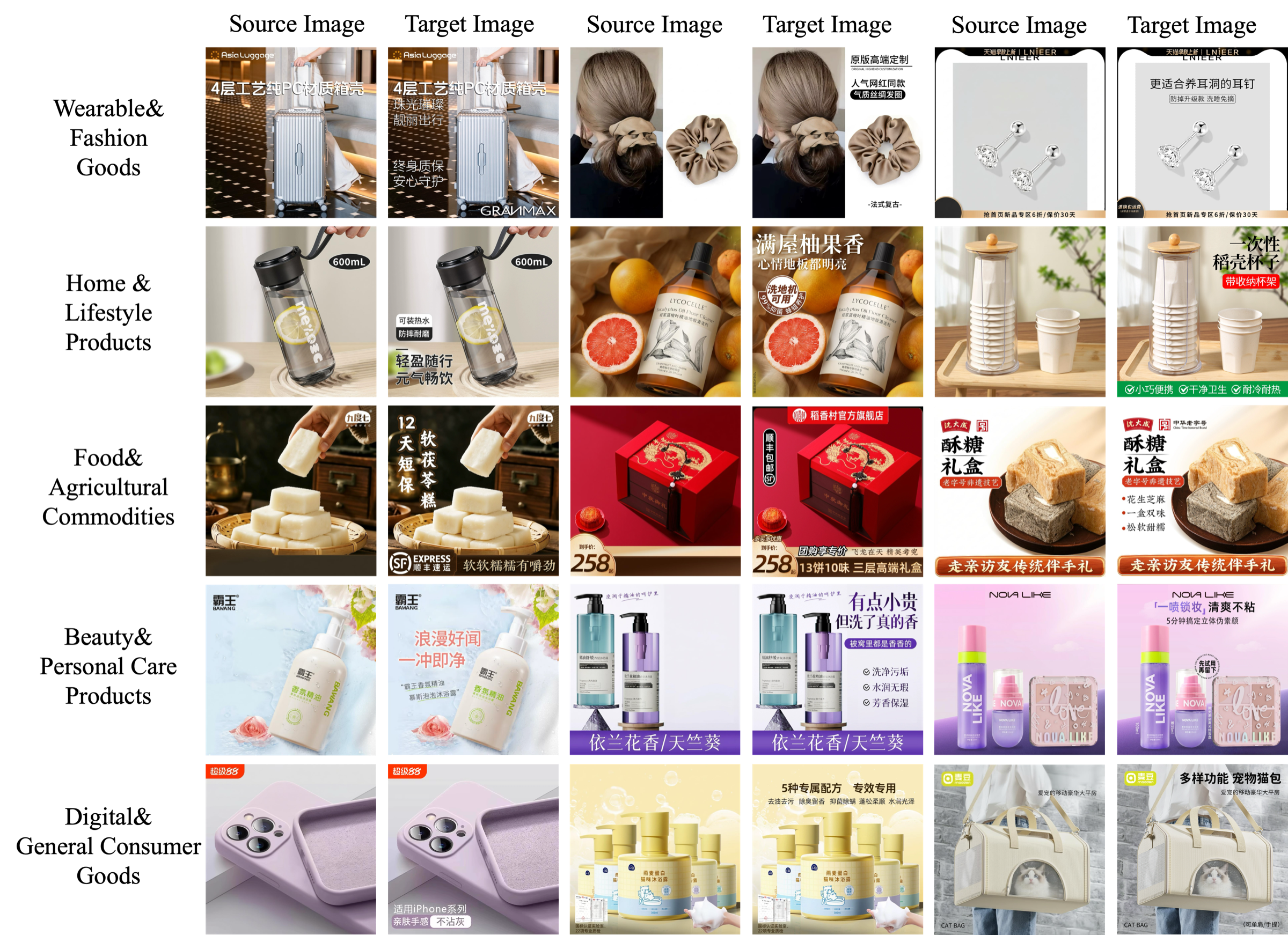}
  \caption{Representative evaluation examples across the five product domains.}
  \label{fig:evaluation-domain-examples}
\end{figure*}

\clearpage

\subsection{A.2 More Experimental Results}
\label{app:more-experimental-results}

\subsubsection{More Results for Text Insertion}
We provide a finer-grained comparison under two complementary OCR normalization protocols. The \textit{Symbol-Invariant} (SI) protocol removes whitespace and punctuation before matching, retaining only Chinese characters, letters, and digits; it therefore measures lexical-content fidelity independently of symbols. The stricter \textit{Symbol-Preserving} (SP) protocol canonicalizes full-width characters and equivalent Chinese/English punctuation, but retains symbols during matching. Both protocols otherwise use the same two-stage assignment, Levenshtein similarity thresholds, extra-OCR penalty, and image-level accuracy definition. For compactness, we abbreviate Qwen-Image-Edit-2511 as \textbf{QIE}, FireRed-Image-Edit-1.0 as \textbf{FireRed}, and our SFT model as \textbf{TR-SFT}; their reward-optimized variants are denoted by the suffix \textbf{-RL}. Tables~\ref{tab:simple-si-results}--\ref{tab:descriptive-sp-results} report results for the simple-template and descriptive-prompt settings.

We additionally compare with PosterMaker, AnyText v2, and GPT-Image-2 (abbreviated as \textbf{GPT-I2}). Unlike the mask-free methods evaluated in our primary setting, PosterMaker and AnyText v2 require an explicit spatial mask as input. To enable a controlled comparison, we use Gemini to generate a semantic description of each pre-edit image and construct the required masks directly from the ground-truth bounding boxes of the target text regions. Thus, these two methods are supplied with explicit localization guidance that is unavailable to the mask-free methods, and their results should be interpreted as a reference under a less restrictive input setting. GPT-Image-2 is evaluated under the same mask-free textual-instruction setting as our method.

\begin{table}[h]
\centering
\footnotesize
\setlength{\tabcolsep}{1.8pt}
\caption{Simple-template prompt under the SI protocol.}
\label{tab:simple-si-results}
\begin{tabular}{lcccccc}
\toprule
Method & Match$\uparrow$ & Partial$\downarrow$ & Miss$\downarrow$ & Extra$\downarrow$ & Score$\uparrow$ & Acc.$\uparrow$ \\
\midrule
PosterMaker& 83.4 & 16.6 & 0.0 & 0 & 0.9685 & 64.5 \\
AnyText v2 & 76.4 & 23.6 & 0.0 & 0 & 0.9552 & 56.0 \\
\spacedhdashline
QIE        & 77.1 & 6.4 & 16.6 & 92  & 0.8361 & 57.5 \\
FireRed    & 85.4 & \textbf{5.6} & 9.1 & \textbf{70} & 0.9109 & 67.0 \\
TR-SFT     & \textbf{89.8} & 6.1 & \textbf{4.1} & 124 & \textbf{0.9536} & \textbf{72.0} \\
\spacedhdashline
GPT-I2     & 96.2 & 3.3 & 0.5 & 22 & 0.9889 & 87.5 \\
\spacedhdashline
QIE-RL     & 91.6 & 7.5 & \textbf{1.0} & \textbf{29} & 0.9764 & 77.0 \\
FireRed-RL & 93.2 & 5.9 & \textbf{1.0} & 32 & \textbf{0.9805} & \textbf{81.0} \\
TR-RL      & \textbf{93.5} & \textbf{4.9} & 1.6 & 45 & 0.9758 & \textbf{81.0} \\
\bottomrule
\end{tabular}
\end{table}

\begin{table}[h]
\centering
\footnotesize
\setlength{\tabcolsep}{1.8pt}
\caption{Simple-template prompt under the SP protocol.}
\label{tab:simple-sp-results}
\begin{tabular}{lcccccc}
\toprule
Method & Match$\uparrow$ & Partial$\downarrow$ & Miss$\downarrow$ & Extra$\downarrow$ & Score$\uparrow$ & Acc.$\uparrow$ \\
\midrule
PosterMaker& 73.4 & 26.6 & 0.0 & 0 & 0.9496 & 48.0 \\
AnyText v2 & 55.2 & 44.8 & 0.0 & 0 & 0.9148 & 34.0 \\
\spacedhdashline
QIE        & 55.8 & 11.3 & 32.9 & 147 & 0.6933 & 35.0 \\
FireRed    & 75.4 & 10.5 & 14.1 & \textbf{106} & 0.8551 & 50.5 \\
TR-SFT     & \textbf{87.1} & \textbf{8.3} & \textbf{4.6} & 135 & \textbf{0.9443} & \textbf{63.5} \\
\spacedhdashline
GPT-I2     & 94.1 & 5.1 & 0.8 & 26 & 0.9824 & 82.5 \\
\spacedhdashline
QIE-RL     & 88.2 & 9.4 & 2.4 & \textbf{37} & 0.9645 & 71.0 \\
FireRed-RL & 91.3 & 7.6 & \textbf{1.1} & 38 & \textbf{0.9755} & 75.0 \\
TR-RL      & \textbf{91.6} & \textbf{6.8} & 1.6 & 49 & 0.9733 & \textbf{75.5} \\
\bottomrule
\end{tabular}
\end{table}

\newpage
\begin{table}[h]
\centering
\footnotesize
\setlength{\tabcolsep}{1.8pt}
\caption{Descriptive prompt under the SI protocol.}
\label{tab:descriptive-si-results}
\begin{tabular}{lcccccc}
\toprule
Method & Match$\uparrow$ & Partial$\downarrow$ & Miss$\downarrow$ & Extra$\downarrow$ & Score$\uparrow$ & Acc.$\uparrow$ \\
\midrule
PosterMaker& 90.0 & 10.0 & 0.0 & 0 & 0.9810 & 69.0 \\
AnyText v2 & 84.2 & 15.8 & 0.0 & 0 & 0.9700 & 49.5 \\
\spacedhdashline
QIE        & 84.7 & 4.2 & 11.1 & 114 & 0.8774 & 50.5 \\
FireRed    & \textbf{91.8} & \textbf{3.7} & \textbf{4.6} & 97 & \textbf{0.9431} & \textbf{71.5} \\
TR-SFT     & 88.8 & 4.6 & 6.7 & \textbf{86} & 0.9256 & 62.0 \\
\spacedhdashline
GPT-I2     & 97.1 & 2.3 & 0.6 & 38 & 0.9900 & 87.0 \\
\spacedhdashline
QIE-RL     & 92.7 & 2.8 & 4.6 & \textbf{89} & 0.9516 & 72.0 \\
FireRed-RL & 94.1 & \textbf{2.6} & 2.8 & 93 & 0.9681 & 79.5 \\
TR-RL      & \textbf{94.4} & 3.4 & \textbf{2.1} & \textbf{89} & \textbf{0.9741} & \textbf{80.0} \\
\bottomrule
\end{tabular}
\end{table}

\begin{table}[h]
\centering
\footnotesize
\setlength{\tabcolsep}{1.8pt}
\caption{Descriptive prompt under the SP protocol.}
\label{tab:descriptive-sp-results}
\begin{tabular}{lcccccc}
\toprule
Method & Match$\uparrow$ & Partial$\downarrow$ & Miss$\downarrow$ & Extra$\downarrow$ & Score$\uparrow$ & Acc.$\uparrow$ \\
\midrule
PosterMaker& 88.0 & 12.0 & 0.0 & 0 & 0.9772 & 63.5 \\
AnyText v2 & 81.6 & 18.4 & 0.0 & 0 & 0.9650 & 44.0 \\
\spacedhdashline
QIE        & 82.0 & 6.7 & 11.3 & 118 & 0.8718 & 45.5 \\
FireRed    & \textbf{89.9} & \textbf{5.2} & \textbf{4.9} & 105 & \textbf{0.9373} & \textbf{66.5} \\
TR-SFT     & 86.1 & 6.7 & 7.2 & \textbf{90} & 0.9157 & 56.5 \\
\spacedhdashline
GPT-I2     & 95.9 & 3.4 & 0.7 & 42 & 0.9868 & 82.0 \\
\spacedhdashline
QIE-RL     & 90.4 & 5.0 & 4.6 & \textbf{92} & 0.9478 & 66.5 \\
FireRed-RL & 92.4 & 4.9 & 2.7 & 96 & 0.9652 & 72.5 \\
TR-RL      & \textbf{92.9} & \textbf{4.7} & \textbf{2.4} & 94 & \textbf{0.9694} & \textbf{73.5} \\
\bottomrule
\end{tabular}
\end{table}

\subsubsection{More Results for Localized Text Replacement}
We further evaluate reward optimization for text replacement. Table~\ref{tab:replacement-reward-results} reports image-level editing accuracy under the same evaluation setting. The comparison consistently uses the corresponding base models and their reward-optimized variants.

\begin{table}[h]
\centering
\footnotesize
\setlength{\tabcolsep}{10pt}
\caption{Additional comparison of reward optimization for text replacement.}
\label{tab:replacement-reward-results}
\begin{tabular}{lc}
\toprule
Method & Acc. (\%)$\uparrow$ \\
\midrule
QIE        & 45.0 \\
FireRed    & 52.0 \\
TR-SFT     & \textbf{77.0} \\
\spacedhdashline
QIE-RL     & 67.0 \\
FireRed-RL & 62.5 \\
TR-RL      & \textbf{85.0} \\
\bottomrule
\end{tabular}
\end{table}

\clearpage
\subsubsection{Qualitative Effectiveness of the Add-Text Reward}
\begin{figure*}[p]
  \centering
  \includegraphics[width=\textwidth,height=0.88\textheight,keepaspectratio]{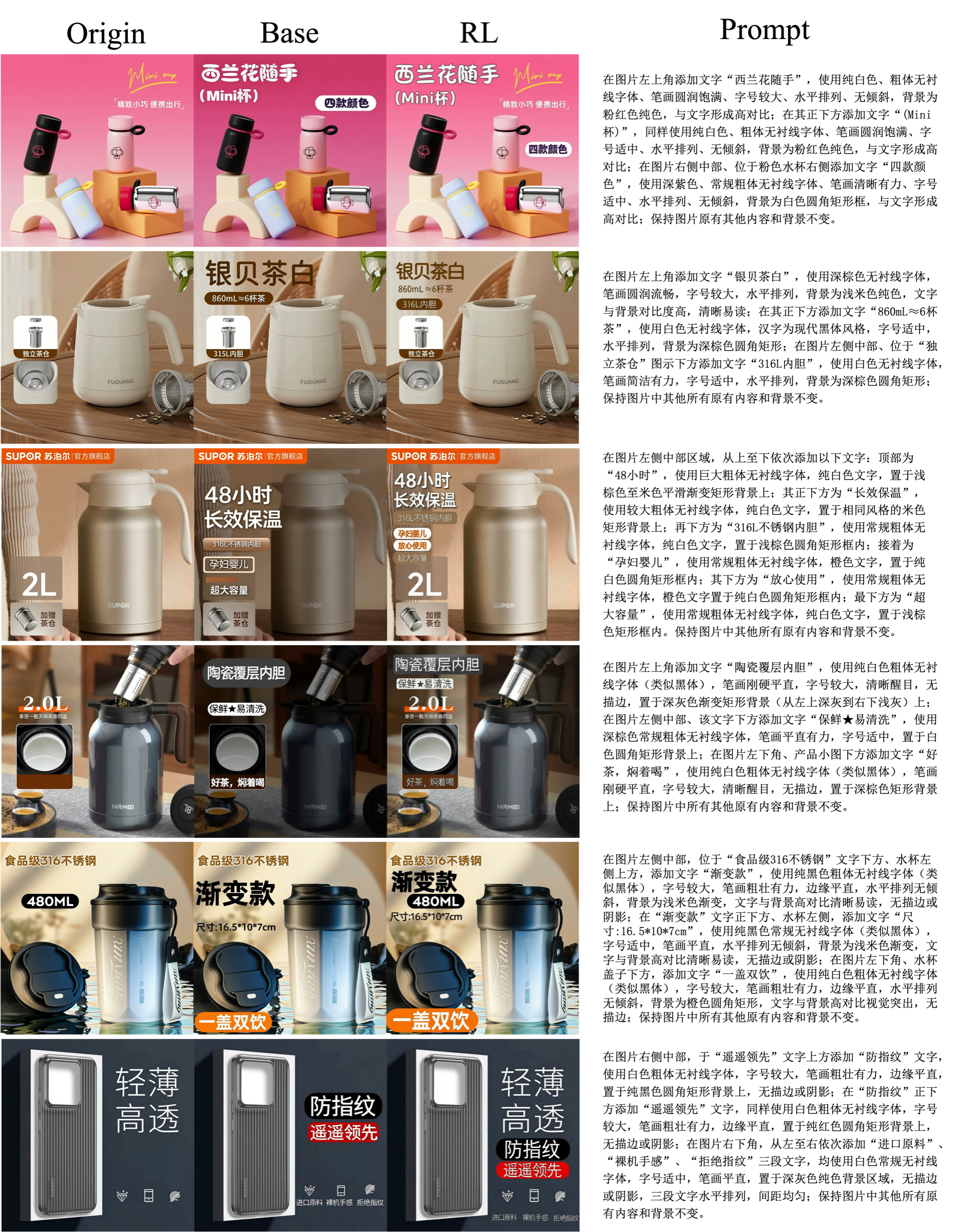}
  \caption{Qualitative comparison before and after applying the add-text reward to Qwen-Image-Edit-2511.}
  \label{fig:qwen-add-text-reward}
\end{figure*}

\begin{figure*}[p]
  \centering
  \includegraphics[width=\textwidth,height=0.88\textheight,keepaspectratio]{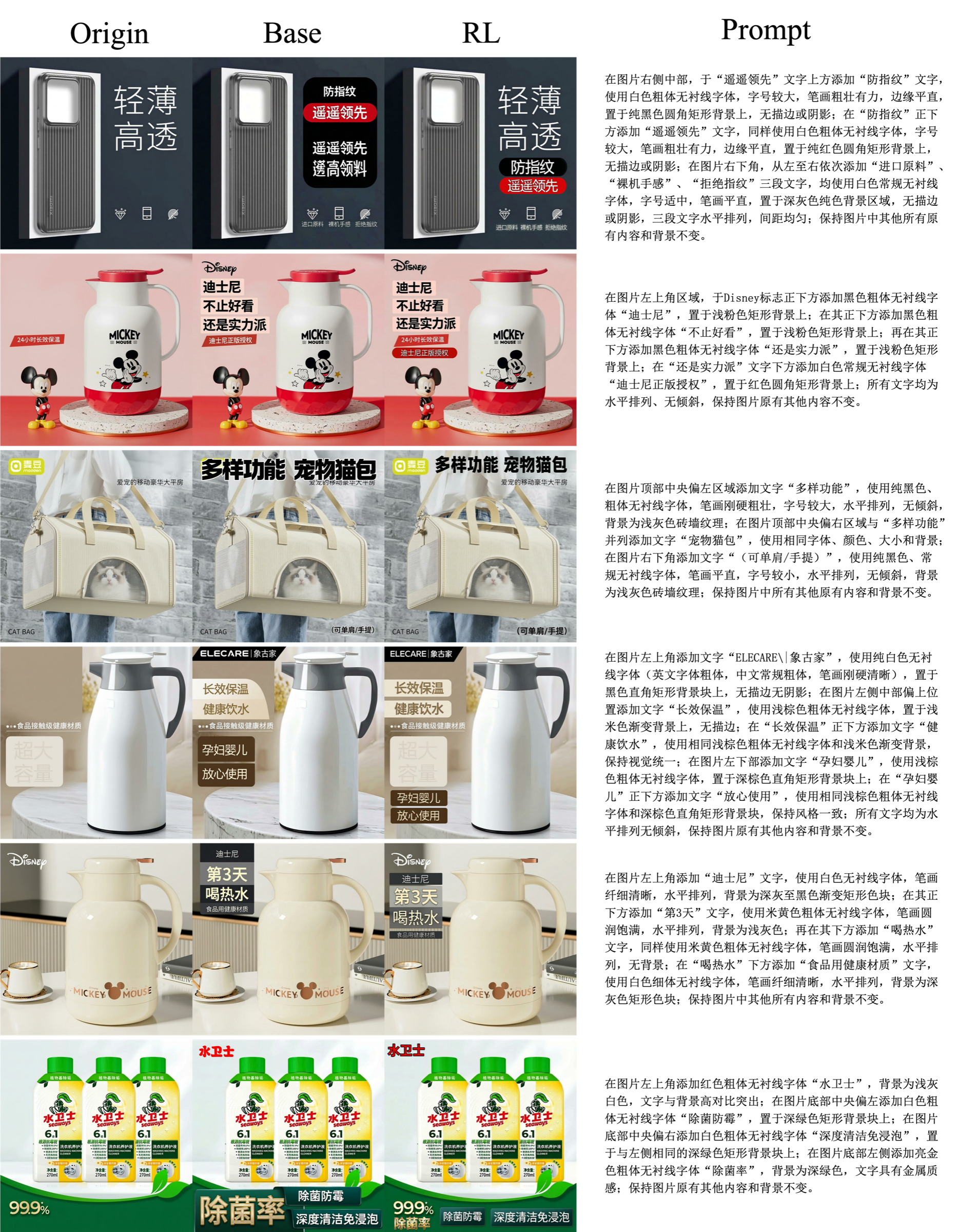}
  \caption{Qualitative comparison before and after applying the add-text reward to FireRed-Image-Edit-1.0.}
  \label{fig:firered-add-text-reward}
\end{figure*}

\begin{figure*}[p]
  \centering
  \includegraphics[width=\textwidth,height=0.88\textheight,keepaspectratio]{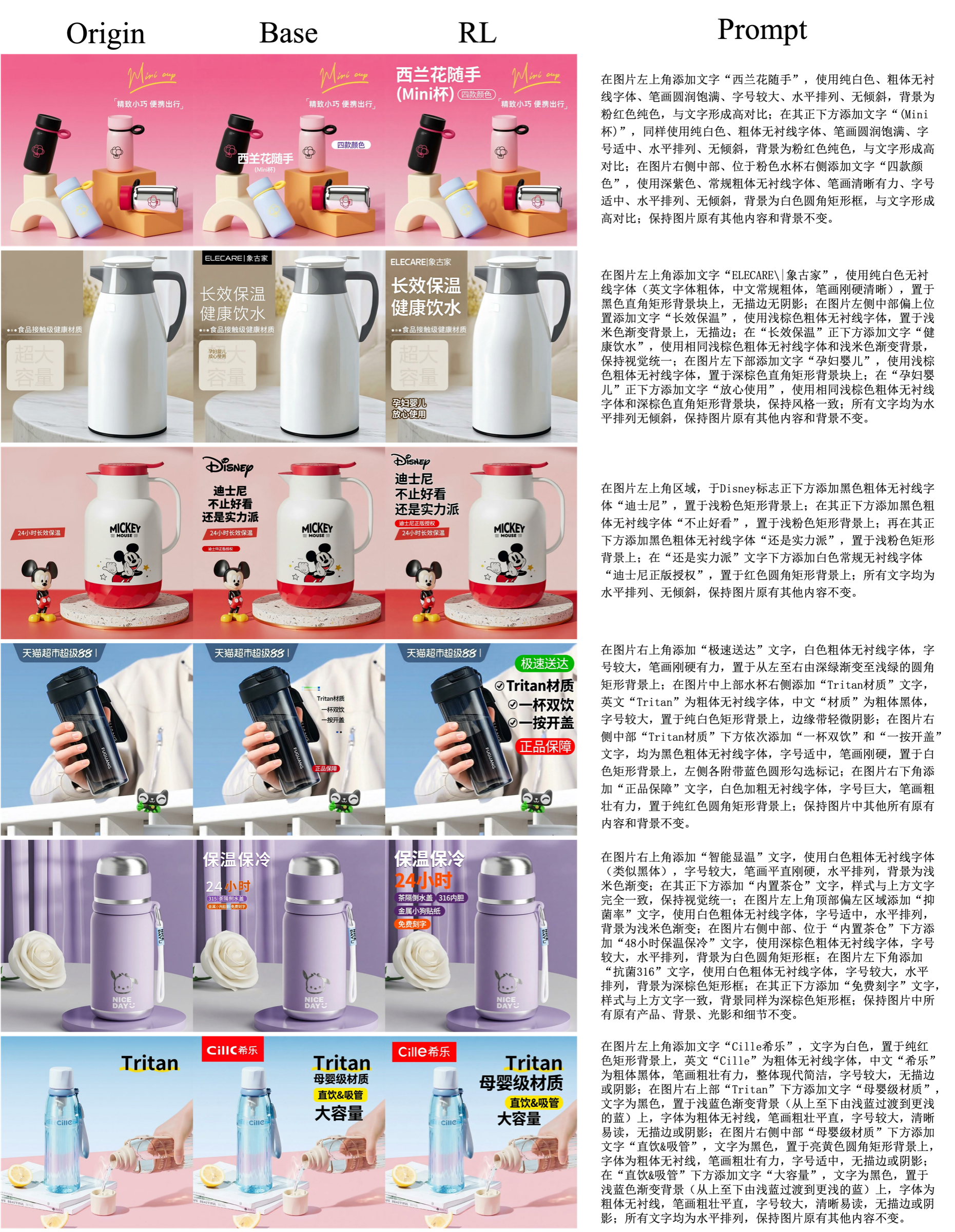}
  \caption{Qualitative comparison of TextRefine before and after reinforcement learning on the add-text task.}
  \label{fig:textrefine-add-text-reward}
\end{figure*}

We further provide qualitative comparisons to illustrate the effectiveness of the add-text reward. Figures~\ref{fig:qwen-add-text-reward}, \ref{fig:firered-add-text-reward}, and~\ref{fig:textrefine-add-text-reward} compare the editing results before and after reward optimization for Qwen-Image-Edit-2511, FireRed-Image-Edit-1.0, and TextRefine, respectively. After reinforcement learning, all three models more reliably insert the requested text while preserving the surrounding product appearance, layout, and background consistency, indicating that the add-text reward directly improves instruction-following fidelity for localized text insertion.

\subsubsection{Qualitative Effectiveness of the Glyph-Level Reward}
\begin{figure*}[p]
  \centering
  \includegraphics[width=\textwidth,height=0.88\textheight,keepaspectratio]{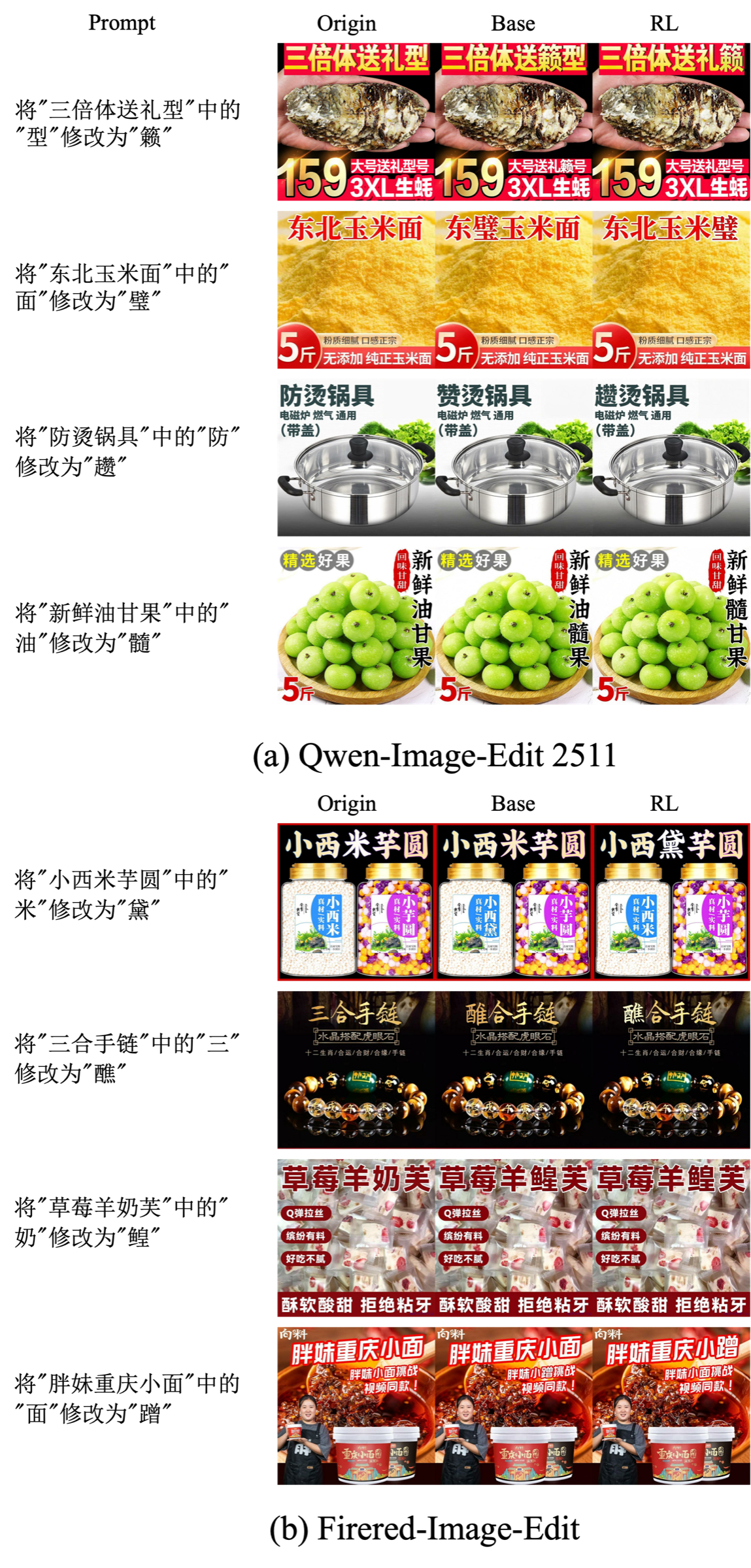}
  \caption{Qualitative comparisons of Qwen-Image-Edit-2511 and FireRed-Image-Edit-1.0 before and after reinforcement learning on the localized text-replacement task.}
  \label{fig:qwen-firered-text-replacement-reward}
\end{figure*}

\begin{figure*}[p]
  \centering
  \includegraphics[width=\textwidth,height=0.88\textheight,keepaspectratio]{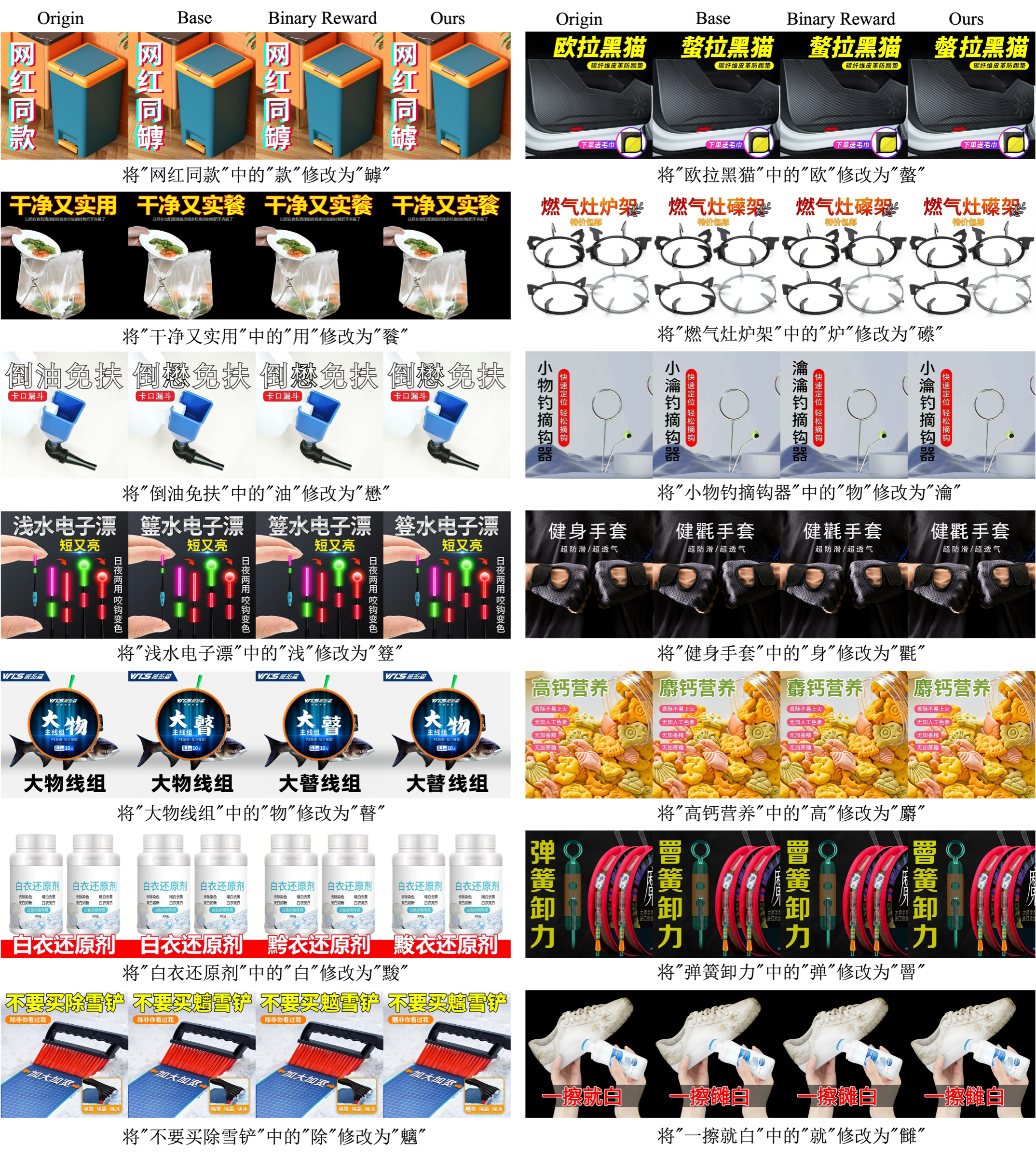}
  \caption{Qualitative comparison of TextRefine before and after reinforcement learning on the localized text-replacement task.}
  \label{fig:textrefine-text-replacement-reward}
\end{figure*}

We additionally visualize the effect of the glyph-level reward on localized text replacement. Figures~\ref{fig:qwen-firered-text-replacement-reward} and~\ref{fig:textrefine-text-replacement-reward} compare results before and after reward optimization for Qwen-Image-Edit-2511 and FireRed-Image-Edit-1.0, and for TextRefine, respectively. Consistent with the quantitative improvements in Table~\ref{tab:replacement-reward-results}, reinforcement learning more reliably replaces the target character with the requested low-frequency glyph while preserving the surrounding typography, product appearance, and poster layout. These examples further show that the target-character CTC-posterior reward provides fine-grained supervision for structurally complex glyphs and reduces missing strokes, malformed components, and confusion among visually similar characters.

\subsubsection{General Image-Editing Evaluation on RedBench}
To examine whether improved text-editing capability compromises general image-editing performance, we further evaluate the models on the FireRed-Image-Edit benchmark, which covers diverse editing operations in both English (RedBench-EN) and Chinese (RedBench-CN). Here, \textbf{QIE} denotes Qwen-Image-Edit-2511, \textbf{FireRed} denotes FireRed-Image-Edit-1.0, and \textbf{TR-RL} denotes our reward-optimized model. As shown in Tables~\ref{tab:redbench-en} and~\ref{tab:redbench-cn}, TR-RL improves the overall score over its QIE initialization in both language settings, indicating that the enhanced text-rendering capability does not sacrifice general image-editing performance.

\newpage
\begin{table}[h]
\centering
\footnotesize
\setlength{\tabcolsep}{4pt}
\caption{General image-editing results on RedBench-EN.}
\label{tab:redbench-en}
\begin{tabular}{lccc:c}
\toprule
Metric & QIE & FireRed & TR-RL & GPT-I2 \\
\midrule
Add        & 4.37 & \textbf{4.52} & 4.45 & 4.70 \\
Adjust     & 3.59 & \textbf{3.93} & 3.61 & 4.12 \\
Background & 3.71 & \textbf{4.23} & 3.92 & 4.48 \\
Beauty     & 2.82 & 2.66 & \textbf{3.03} & 4.00 \\
Color      & 3.84 & \textbf{4.12} & 3.83 & 4.37 \\
Compose    & 3.46 & 3.49 & \textbf{3.60} & 4.07 \\
Extract    & 2.29 & 2.35 & \textbf{3.58} & 3.36 \\
Low-level  & 2.68 & 3.17 & \textbf{3.50} & 3.55 \\
Motion     & 4.61 & 4.62 & \textbf{4.71} & 4.78 \\
Portrait   & \textbf{4.28} & 3.63 & 3.90 & 4.64 \\
Remove     & 4.28 & 4.22 & \textbf{4.33} & 4.27 \\
Replace    & 4.36 & 4.42 & \textbf{4.45} & 4.60 \\
Stylize    & 4.76 & \textbf{4.80} & 4.76 & 4.99 \\
Text       & 3.69 & 3.72 & \textbf{3.83} & 4.30 \\
Viewpoint  & 2.61 & \textbf{2.75} & 2.66 & 3.02 \\
\midrule
Final Score & 3.69 & 3.78 & \textbf{3.88} & 4.22 \\
\bottomrule
\end{tabular}
\end{table}

\begin{table}[h]
\centering
\footnotesize
\setlength{\tabcolsep}{4pt}
\caption{General image-editing results on RedBench-CN.}
\label{tab:redbench-cn}
\begin{tabular}{lccc:c}
\toprule
Metric & QIE & FireRed & TR-RL & GPT-I2 \\
\midrule
Add        & 4.43 & \textbf{4.48} & 4.42 & 4.67 \\
Adjust     & 3.44 & \textbf{3.90} & 3.67 & 4.07 \\
Background & 3.94 & \textbf{4.35} & 4.02 & 4.45 \\
Beauty     & 2.79 & 2.52 & \textbf{3.02} & 3.91 \\
Color      & 3.71 & \textbf{3.94} & 3.86 & 4.40 \\
Compose    & 3.35 & 3.63 & \textbf{3.69} & 4.11 \\
Extract    & 2.26 & 2.55 & \textbf{3.42} & 3.37 \\
Low-level  & 2.59 & 3.07 & \textbf{3.55} & 3.39 \\
Motion     & \textbf{4.59} & 4.58 & 4.58 & 4.88 \\
Portrait   & 3.54 & 3.82 & \textbf{3.85} & 4.57 \\
Remove     & 4.15 & 4.24 & \textbf{4.30} & 4.36 \\
Replace    & 4.38 & \textbf{4.50} & 4.40 & 4.66 \\
Stylize    & 4.69 & 4.73 & \textbf{4.83} & 4.99 \\
Text       & 3.82 & 3.67 & \textbf{3.83} & 4.33 \\
Viewpoint  & 2.40 & 2.73 & \textbf{3.20} & 3.45 \\
\midrule
Final Score & 3.61 & 3.78 & \textbf{3.91} & 4.24 \\
\bottomrule
\end{tabular}
\end{table}

\clearpage
\clearpage

\clearpage

\clearpage

\clearpage

\clearpage

% Add subsequent appendix sections here.

% Uncomment this after adding citations to the appendix.
% \bibliography{cofirender_references}

\end{document}